\pdfoutput=1

\documentclass[11pt]{article}

\usepackage[dvipsnames]{xcolor}

\usepackage{placeins}

\PassOptionsToPackage{sort}{natbib}
\PassOptionsToPackage{hyphens}{url}\usepackage{hyperref}

\usepackage[final]{acl}

\usepackage{times}
\usepackage{latexsym}
\usepackage{booktabs}
\usepackage{fancyvrb}
\usepackage{multirow}
\usepackage[normalem]{ulem} 
\usepackage{dblfloatfix}
\usepackage{float}

\usepackage[T1]{fontenc}

\usepackage[utf8]{inputenc}

\usepackage{microtype}

\usepackage{inconsolata}

\usepackage{graphicx}
\usepackage{amsmath}
\usepackage{amssymb}
\usepackage{float}
\usepackage{hyperref}
\let\vec\mathbf

\usepackage{subcaption}

\title{All for One: LLMs Solve Mental Math at the Last Token \\With Information Transferred From Other Tokens}

\author{
 \textbf{Siddarth Mamidanna\textsuperscript{1}},
 \textbf{Daking Rai\textsuperscript{2}},
 \textbf{Ziyu Yao\textsuperscript{2}},
 \textbf{Yilun Zhou\textsuperscript{3}}
\\
\\
 \textsuperscript{1}University of California, Santa Cruz,
 \textsuperscript{2}George Mason University,
 \textsuperscript{3}Datadog AI Research
\\\\
 Correspondence to: \url{spmamida@ucsc.edu}
\\
\url{https://github.com/siddarth-pm/all-for-one}
}

\newcommand{\modelname}{\textsc{AF1}}
\newcommand{\lwait}{L_\mathrm{wait}}
\newcommand{\ltrans}{L_\mathrm{transfer}}
\newcommand{\colora}[1]{\colorbox{gray!30}{\strut \textcolor{black}{\texttt{#1}}}}
\newcommand{\colorb}[1]{\colorbox{gray!50}{\strut \textcolor{black}{\texttt{#1}}}}

\definecolor{GroupA}{RGB}{31,119,180}   
\definecolor{GroupB}{RGB}{255,127,14}   
\definecolor{GroupC}{RGB}{44,160,44}    
\definecolor{GroupD}{RGB}{214,39,40}    

\newcommand{\groupa}[1]{\colorbox{GroupA!20}{#1}}
\newcommand{\groupb}[1]{\colorbox{GroupB!20}{#1}}
\newcommand{\groupc}[1]{\colorbox{GroupC!20}{#1}}
\newcommand{\groupd}[1]{\colorbox{GroupD!20}{#1}}

\newcommand{\reviseadd}[1]{#1}

\newtoggle{comments}
\toggletrue{comments}

\iftoggle{comments}{
    \newcommand{\yilun}[1]{\textcolor{blue}{(Yilun: #1)}}
    \newcommand{\ziyu}[1]{\textcolor{orange}{(Ziyu: #1)}}
    \newcommand{\daking}[1]{#1}
    \newcommand{\sid}[1]{\textcolor{ForestGreen}{(Sid: #1)}}
}{
    \newcommand{\yilun}[1]{}
    \newcommand{\ziyu}[1]{}
    \newcommand{\daking}[1]{}
    \newcommand{\sid}[1]{}
}

\makeatletter
\newcommand\footnoteref[1]{\protected@xdef\@thefnmark{\ref{#1}}\@footnotemark}
\makeatother

\begin{document}
\maketitle
\begin{abstract}

Large language models (LLMs) demonstrate proficiency across numerous computational tasks, yet their inner workings remain unclear. In theory, the combination of causal self-attention and multilayer perceptron layers allows every token to access and compute information based on all preceding tokens. In practice, to what extent are such operations present? In this paper, on mental math tasks (i.e., direct math calculation via next-token prediction without explicit reasoning), we investigate this question in three steps: inhibiting input-specific token computations in the initial layers, restricting the routes of information transfer across token positions in the next few layers, and forcing all computation to happen at the last token in the remaining layers. With two proposed techniques, Context-Aware Mean Ablation (CAMA) and Attention-Based Peeking (ABP), we identify an All-for-One subgraph (\modelname{}) with high accuracy on a wide variety of mental math tasks, where meaningful computation occurs very late (in terms of layer depth) and only at the last token, which receives information of other tokens in few specific middle layers. Experiments \reviseadd{on a variety of models and arithmetic expressions} show that this subgraph is sufficient and necessary for high model performance, transfers across different models, and works on a variety of input styles. Ablations on different CAMA and ABP alternatives reveal their unique advantages over other methods, which may be of independent interest.

\end{abstract}

\section{Introduction}

\begin{figure}[h!]
    \centering
    \includegraphics[width=1\linewidth]{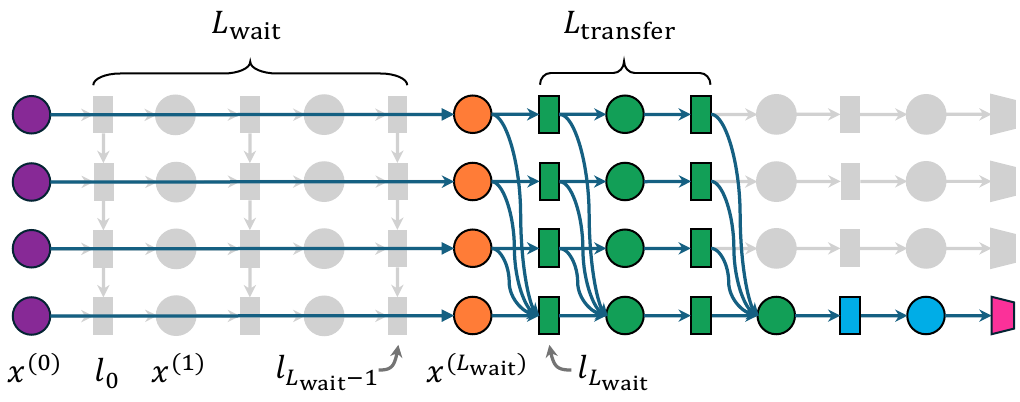}
    \caption{The full \modelname{} subgraph consists of three stages. 
    \daking{First, input-specific computation is suppressed} where the \textcolor{Plum}{input embeddings} skip the first $\lwait$ layers with \textcolor{gray}{context-aware mean ablation (CAMA)}. 
    Then, the \textcolor{Orange}{resulting activations $x^{(\lwait)}$} pass through $\ltrans$ layers of \textcolor{ForestGreen}{attention-based peeking (ABP)} where the only cross-token attentions are those from the last token to preceding ones. Last, for the remaining \textcolor{Cerulean}{remaining ABP layers}, the last token only attends to itself without any cross-token attention to finish computation, ending in the \textcolor{Magenta}{outputs}. In this diagram, $\lwait=3$ and $\ltrans=2$.}
    \label{fig:1}
\end{figure}

Large language models (LLMs) perform well on a multitude of computational tasks, and one of the biggest contributing factors is the transformer architecture~\cite{vaswani2017attention}. Unlike its recurrent neural network (RNN) predecessor, a transformer allows for any token to immediately access all preceding tokens for information transfer via self-attention and enables each token to carry out its independent computation in parallel via multilayer perceptron (MLP). However, it is not clear to what extent these operations are actually happening. 

\reviseadd{In this paper, we focus on ``mental math'' tasks of two and three operands (i.e., arithmetic problems that can be solved with a one token response without explicit chain-of-thought reasoning by the model, such as $42 + 20 - 15$) and investigate the “least amount of computation” that still allows the model to perform well.} Specifically, we ask the following questions. First, although each token can access all preceding tokens at every layer, is such access actually executed from the beginning? Second, do all tokens need to perform computation, or is the last-token computation sufficient, given that the next token is predicted from its final residual stream representation? Last, does the last-token computation need access to all other tokens in all layers, or can it function after a (short) period of information transfer?

To answer these questions, we progressively modify the vanilla transformer architecture with two techniques, Context-Aware Mean Ablation (CAMA) and Attention-Based Peeking (ABP), and get a surprisingly sparse subgraph that performs well on a wide variety of mental math prompts, with three stages (Fig.~\ref{fig:1}). First, \daking{in the early layers,} all tokens wait to access other tokens and instead perform task-general computation \daking{(e.g., understanding that the next-token prediction requires three-operand arithmetic)} without input-specific information \daking{(e.g., the numerical value of the first operand)} from other input tokens. Second, \daking{in a few middle layers,} all tokens transfer their information to the last token. \daking{Finally, in the remaining layers,} the last token continues the computation to yield the next token prediction. Since input-specific computation is only computed at the last token with information transferred from other tokens, we call this computation subgraph All-for-One (\modelname{}). We use $\lwait$ and $\ltrans$ to represent the number of layers in the first two stages and study this family of subgraphs extensively. Notably, for Llama-3-8B \reviseadd{and Llama-3.1-8B}~\citep{grattafiori2024llama}, and, to a lesser extent, Pythia~\citep{biderman2023pythia} and GPT-J~\citep{gpt-j}, the first stage can be extended quite long, to nearly half of all layers, while the second stage only needs as few as two layers to recover high performance.

This paper makes three main contributions. First, the fact that \modelname{} works well on a two-hop arithmetic task (e.g., $A+B+C$) suggests a lack of compositionality at different token positions (e.g., computing $A+B$ in initial layers, storing it in the token $B$ and then adding $C$ in later layers). Second, the short information transfer period may not be unique to arithmetic, possibly implying that the token residual streams spend most time computing rather than communicating. Last, an ablation study on CAMA and ABP shows that different alternative designs fail to uncover this subgraph, making them potentially useful new tools for investigating other LLM behaviors as well.

\section{Related Work}
\label{sec:related-work}
\textbf{Mechanistic Interpretability }
Mechanistic interpretability (MI) seeks to reverse-engineer the internal mechanisms of LLM behaviors by analyzing their weights and activations~\citep{rai2024practical, bereska2024mechanistic, elhage2021mathematical}. Recent MI work has shed light on a range of LLM capabilities, including in-context learning~\citep{elhage2021mathematical, hendel2023context, ren2024identifying}, reasoning~\citep{stolfo2023mechanistic, rai2024investigation, dutta2024think, biran2024hopping, nikankin2024arithmetic}, and factual recall~\citep{geva2023dissecting, chughtai2024summing, meng2022locating, hernandez2023linearity}. Building on these advances, we investigate how LLMs perform arithmetic reasoning. In addition, prior MI studies have introduced a range of investigative techniques, including ablation, activation patching, and logit lens~\citep{li2024optimal, goldowsky2023localizing, chan2022causal, wang2022interpretability, meng2022locating, nostalgebraist2020blog}. We employ logit lens in our study and also introduce two new methods: CAMA, an ablation approach inspired by mean ablation, and ABP, a technique for enforcing selective information transfer through attention. \reviseadd{Relatedly, \citet{haklay2025positionaware} introduced a position-aware circuit method with similarities to our ABP technique, highlighting the position-dependent nuances that LLMs employ.}

\noindent \textbf{Interpreting LLMs in Arithmetic Reasoning Tasks }
Several prior studies have examined the internal mechanisms of LLMs to understand how they perform arithmetic reasoning~\citep{rai2024investigation, wu2023interpretability, nanda2023progress, maltoni2024arithmetic}. \citet{zhou2024pre} observed that LLM leverages Fourier space features to perform addition. Most relevant to our work, \citet{stolfo2023mechanistic} and \citet{zhang2024interpreting} mapped out the general structure of arithmetic circuits in LLMs, detailing how operands and operators are processed, how information is transferred across layers, and how results are ultimately computed. Similarly, \citet{nikankin2024arithmetic} identified circuits responsible for arithmetic reasoning, which suggests that LLMs rely on a collection of heuristics, each effective over a limited input distribution, rather than a single monolithic algorithm. However, their analysis is restricted to a two-operand template of \texttt{A$\circ$B=}. In contrast, we identify a general-purpose subgraph that faithfully handles both two- and three-operand arithmetic tasks, including both symbolic (e.g., ``$3 + 4 + 5$'') and verbal (e.g., ``The sum of 3 and 4 is'') formulations. 

\noindent \textbf{Interpreting LLMs in Implicit Reasoning }
Several prior studies have investigated how LLMs reason implicitly over parametric knowledge~\citep{yang-etal-2024-large-language-models, sakarvadia-etal-2023-memory, li-etal-2024-understanding}. However, they do not track how information flows across different token positions and layers for multi-hop reasoning. More relevant to our study, \citet{biran2024hopping} and \citep{wang2024grokked} studied multi-hop factual recall and showed that LLMs resolve single-hop answers in earlier layers and propagate the result to final token positions in middle layers, with second-hop factual recall occurring in later layers. We build on this line of work to study the multi-hop reasoning for the three-operand task, but with contradictory findings suggesting a lack of such inter-layer hopping behavior. \reviseadd{Additionally, \citet{csordas2025efficientdepth} observed non-compositional tendancies in models similar to ours, highlighting how LLMs often do not rely on systematic compositional reasoning across novel task formulations.} 

\section{Methods}

\subsection{LLM Architecture Notation}

To standardize presentation, we use the following notations to describe the LLM architecture (see Fig.~\ref{fig:1}). Let $\vec x=\{x_1, ..., x_T\}$ be an input sequence of $T$ tokens. Use $x_t^{(0)}$ to denote the original (token and positional) input embedding for token $x_t$. For $L$ layers, layer $l\in\{0, ..., L-1\}$ takes in $x_t^{(l)}$ and computes $x_t^{(l+1)}$, resulting in the sequence of $x_t^{(1)}, x_t^{(2)}, ..., x_t^{(L)}$ via self-attention and multi-layer perceptron (MLP), which constitute the residual stream.\footnote{We use 1-based indexing for token, and 0-based indexing for layer (and attention head), following common practice.} For model $m$, we write $m(\vec x, t, l)$ as the function that computes $x_t^{(l)}$ by the first $l$ layers. 

\subsection{From Transformer to \modelname{}}
\label{sec:discovery-method}
We develop the \modelname{} subgraph for Llama-3-8B with respect to the $A+B+C$ task, where $A$, $B$, and $C$ are numerical tokens. First, we progressively replaced the first $\lwait$ layers of the full transformer with a ``waiting'' period in which each token can compute independently but not access any other tokens. These computations are task-general, such as understanding numerical tokens \reviseadd{and arithmetic structure}, rather than input-specific, such as performing the operation in the input. We continue this replacement until performance significantly drops, at which point we conclude that information transfer is necessary. We propose a novel waiting mechanism in Sec.~\ref{sec:CAMA}. 

In the second phase, with the waiting period in place, we modify token attentions in all subsequent layers such that the last token $x_T$ can attend to all tokens, while other tokens $x_1, ..., x_{T-1}$ can only attend to themselves.\footnote{\label{footnote:bos}For technical reasons discussed in Sec.~\ref{sec:abp}, we also allow an additional attention to the BOS token.} Despite this drastic attention pruning, performance on the task remains high, matching the full-computation case. The attention pruning is detailed in Sec.~\ref{sec:abp}.

Finally, \reviseadd{with the previous attention pruning already in place}, we only allow $x_T$ to attend to (and hence receive information from) all tokens during the first $\ltrans$ layers after the waiting, for $\ltrans \in\{0, 1, ..., \leq L-\lwait\}$. In the remaining layers, we force it to only attend to itself,\footnoteref{footnote:bos} similar to all other tokens in these layers. These $\ltrans$ layers are thus sufficient for transferring all input-specific information to $x_T$, whose final activation $x_T^{(L)}$ is used by later layers to predict the next token.


We present the detailed results of these three steps for Llama-3-8B on the task of $A+B+C$ in Sec.~\ref{sec:discovery-result}. Qualitatively, we observe that $\lwait$ can be quite long, to almost half of the total number of layers, while $\ltrans$ can be as small as 2, suggesting that a very brief burst of information transfer is sufficient for this arithmetic task. More broadly, although self-attention allows for a quadratic number of information transfer routes, the sufficiency of only a linear number of such routes echoes the success of linear attention transformers \citep{katharopoulos2020transformers}.


\subsection{Token Waiting with Context-Aware Mean Ablation (CAMA)}\label{sec:CAMA}

The transformer architecture immediately gathers and distributes information from and to all tokens starting from the first self-attention layer.\footnote{Technically, the causal attention means that a token can only receive information from its preceding tokens but we ignore this detail for ease of presentation.} However, such information fusion may not be present, especially in early layers, which more likely focus on low-level token features \citep{tenney2019bert}. 

There are many alternative approaches to test for the existence of this waiting period, such as to copy the input embedding directly to the output of the waiting layers, or to allow each token to only attend to itself in these layers. However, as Sec.~\ref{sec:alternative-cama-abp-designs} shows, none of them preserve model performance with a long waiting period, most likely because they lead to out-of-distribution representations. Furthermore, these layers may also perform task-general computation, such as processing a number token to ``understand'' its numerical value. 


In this paper, we propose \textbf{context-aware mean ablation} (CAMA),\footnote{Fun fact: Cama is a hybrid between a Camel and a Llama.} a more ``in-distribution'' ablation approach which is aware of and tailored to the input context distribution and also allows for such task-general computation. Given the distribution $\mathbb P(\vec x)$ over all input sequences, making $x_t$ wait for $\lwait$ layers (i.e., layers 0 to $\lwait-1$) under the CAMA means replacing $x_t^{(\lwait)}$ with 
\begin{align}
    \tilde x_t^{(\lwait)}=\mathbb{E}_{\vec{x}'\sim \mathbb P(\vec{x}\mid x_t)}
    \!\bigl[\,m(\vec{x}',\,t,\,\lwait)\bigr]. \label{eq:cama-definition}
\end{align}

As CAMA replaces the true $x_t^{(\lwait)}$ with the expected representation at the same position over all inputs conditioned on $t$-th token being $x_t$, it preserves the general effect of the context on the representation and allows for task-general computation, while erasing any input-specific information carried by the particular sequence $\vec x$ (other than that of $x_t$). In other words, given the input distribution, \emph{the CAMA value of a specific token yields no new information about other tokens in the input}. 

As we show in Sec.~\ref{sec:alternative-cama-abp-designs}, the CAMA formulation is crucial to uncover the minimal \modelname{} circuit proposed in this paper, with alternative designs being infeasible, suggesting broader applicability to other tasks. In addition, it may be generalizable to ablating out information fusion that does not start in the first layer, which we leave to future work. 


\subsection{Selective Information Transfer with Attention-Based Peeking (ABP)}
\label{sec:abp}
CAMA enforces \emph{waiting} of a token by using context-aware ablation to block a token from accessing input-specific information from other tokens, but it must be applied from the first layer and runs continuously. Here, we introduce an alternative mechanism to control information access at arbitrary layer(s), called \textbf{attention-based peeking} (ABP). It is implemented by modifying the attention mask so that, in the target layer, each query position is allowed to attend only to (or ``peek at'') a subset of previous key positions. 

For a (query) token $x_t$, we use $K_t\subseteq \{1, ..., t\}$ to denote the index set of (key) tokens whose information we want to transfer to $x_t$. Let $M \in \mathbb{R}^{T \times T}$ be the (pre-softmax) attention matrix for the entire sequence $\vec x$ and $K_1, ..., K_T$ be the peeking index sets. We replace each $M_{q,k}$ with $-\infty$ if $k \notin K_q$, so the softmax zeroes out the attention to any keys not in the peek set.

While ABP can be implemented to attend to any subset of (preceding) tokens, in this paper, we consider two specific cases, \reviseadd{\textbf{full-peeking}}, $K_t=\{1, ..., t\}$ where $x_t$ attends to all tokens (and recovers the standard causal attention) and \reviseadd{\textbf{self-peeking}}, $K_t=\{t\}$ where $x_t$ attends only to itself. 





\noindent\textbf{First token attention.}  
A common quirk of attention patterns is the attention sink phenomenon~\citep{xiao2023efficient, cancedda2024spectral}, where tokens strongly attend to the first one, which is often the special <BOS> token. In our experiments, we find that removing the attention to <BOS> is indeed devastating to model performance. Thus, we always keep it with $K_t\leftarrow K_t\cup \{1\}$.

\begin{table}[!t]
    \centering
    \vspace{0.07in}
    \resizebox{\columnwidth}{!}{
    \begin{tabular}{lll}\toprule
    Input template & Tokenization & Operator $\circ$\\\midrule
    \texttt{``<BOS>A$\circ$B=''} & \colora{<BOS>}\colorb{A}\colora{$\circ$}\colorb{B}\colora{=} & $+, -, *, /$\\\midrule
    \texttt{``<BOS>A $\circ$ B $\circ$ C = ''} & 
    \colora{<BOS>}\colorb{A}\colora{\textvisiblespace$\circ$}\colorb{\textvisiblespace}\colora{B}\colorb{\textvisiblespace$\circ$}\colora{\textvisiblespace}\colorb{C}\colora{\textvisiblespace=}\colorb{\textvisiblespace} & $+, -$\\\bottomrule
    \end{tabular}
    }
    \caption{The input templates along with their tokenization (space represented by \textvisiblespace). 
    \daking{$A$, $B$, and $C$ are randomly selected from $\{0, 1, ..., 100\}$ with an additional constraint that the answer from selected input is an integer in the range $\{0, 1, ..., 999\}$.}
    The first template includes spaces around the two operators (which can be different) and the second template does not contain any spaces.}
    \label{tab:task-template}
\end{table}

\section{Experiments}
\label{sec:experiment}

After introducing the experiment setup in Sec.~\ref{subsec:setup}, we present the results of our \modelname{} subgraph discovery process in Sec.~\ref{sec:discovery-result} for Llama-3-8B on the $A+B+C$ task. Then in subsequent sections, we study various properties of this subgraph to demonstrate its generality.

\subsection{Experiment Setup}
\label{subsec:setup}


We consider both two- and three-operand arithmetic, with task templates and their tokenization summarized in Tab.~\ref{tab:task-template}. Each operand is a randomly sampled integer from 0 to 100, subject to the additional constraint that the final answer is an integer from 0 to 999. \reviseadd{The CAMA implementation for these tasks can be found in App.~\ref{app:cama-expanded}.}

We mainly study Llama-3-8B and Llama-3.1-8B, but, to study cross-model generalization, also consider two earlier models, Pythia-6.9B and GPT-J-6B. Since the latter two have very poor three-operand performance, we only study two-operand tasks for them. Tab.~\ref{tab:raw-accuracy} presents the raw model accuracy on randomly sampled legal inputs for all tasks and models.

As model-task accuracy can vary significantly, we use the \textbf{faithfulness} of a subgraph $s$ to measure its performance, defined as its accuracy on prompts $(x, y)$ for which the full model $m$ is correct:
\begin{align}
    \mathrm{faith.}(s) = \mathbb E_{x, y}[s(x)=y|m(x)=y].
\end{align}

To calculate CAMA values for each token, we draw input $\vec x'$ with different operand values but of the same task. In other words, we randomize all operands (other than the token itself, if applicable) while fixing the operator(s), along with the equality sign and any space tokens, when present.


\begin{table}[!t]
    \centering
    \resizebox{\columnwidth}{!}{
    \begin{tabular}{llr|llr}\toprule
        Operation & Model & Raw Acc. & Operation & Model & Raw Acc.\\\midrule
        $A+B+C$ & Llama-3-8B & 0.994 & $A+B$ & Llama-3-8B & 0.962\\
        $A+B-C$ & Llama-3-8B & 0.932 & $A-B$ & Llama-3-8B & 0.899 \\
        $A-B+C$ & Llama-3-8B & 0.296 & $A\times B$ & Llama-3-8B & 0.937 \\
        $A-B-C$ & Llama-3-8B & 0.748 & $A\div B$ & Llama-3-8B & 0.966 \\\midrule
        $A+B+C$ & Llama-3.1-8B & 0.998 & $A+B$ & Llama-3.1-8B & 0.953\\
        $A+B-C$ & Llama-3.1-8B & 0.956 & $A-B$ & Llama-3.1-8B & 0.518 \\
        $A-B-C$ & Llama-3.1-8B & 0.956 & $A\div B$ & Llama-3.1-8B & 0.814 \\
        $A-B+C$ & Llama-3.1-8B & 0.977 & $A\times B$ & Llama-3.1-8B & 0.737 \\\midrule
        $A+B$ & Pythia & 0.584 & $A+B$ & GPT-J & 0.559\\
        $A-B$ & Pythia & 0.226 & $A-B$ & GPT-J & 0.357\\
        $A\times B$ & Pythia & 0.131 & $A\times B$ & GPT-J & 0.316\\
        $A\div B$ & Pythia & 0.220 & $A\div B$ & GPT-J & 0.313\\\bottomrule
    \end{tabular}
    }
    \caption{Raw accuracy of various models for different math operations.} 
    \label{tab:raw-accuracy}
\end{table}


\subsection{\modelname{} Subgraph Discovery Result}
\label{sec:discovery-result}

Sec.~\ref{sec:discovery-method} details the discovery process of the ``All-for-One'' (AF1) subgraph for the three-operand task \(A + B + C\) via a three-phase ablation and peeking procedure. Here, we present the results for Llama-3-8B. 

\begin{figure}[!b]
    \centering
    \includegraphics[width=\columnwidth]{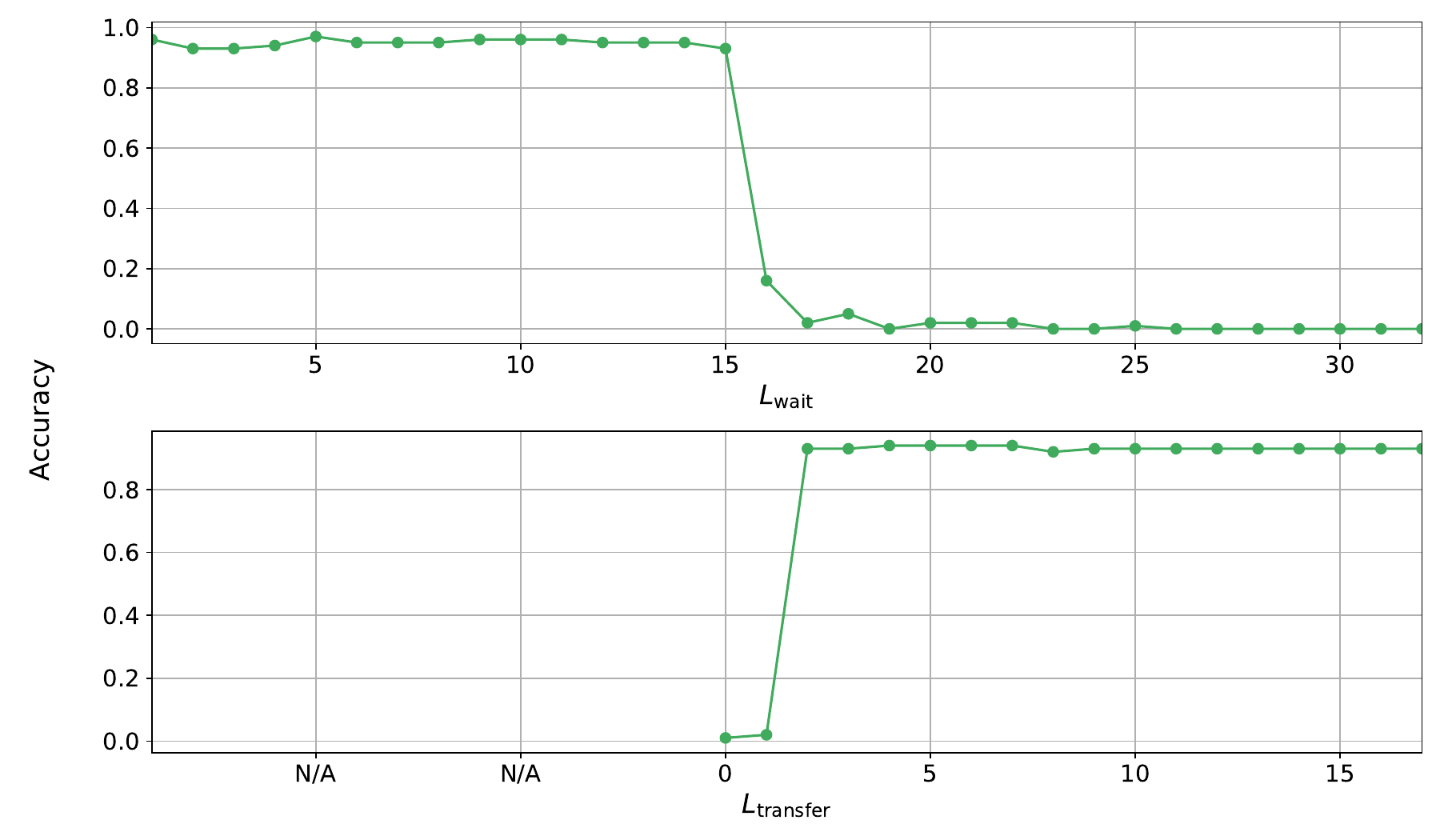}
    \caption{Top: Llama-3-8B performance after making tokens to wait for the first $\lwait$ layers with CAMA, as a function of $\lwait$. \reviseadd{Bottom: for $\lwait=15$ and self-peeking on all non-last tokens, performance of model with last token full-peeking in the next $\ltrans$ layers and self-peeking afterward, as a function of $\ltrans$.}}
    \label{fig:discovery}
\end{figure}

Fig.~\ref{fig:discovery} (top) tracks the faithfulness when the first $\lwait$ layers of the full transformer are replaced with Context-Aware Mean Ablation (CAMA). Faithfulness remains high for \(\lwait \le 15\) and then collapses sharply at \(\lwait = 16\), indicating that no cross-token interaction before layer \reviseadd{index} 14 is necessary, but information transfer at layer \reviseadd{index} 15 is critical \reviseadd{(recall that layer index is 0-based)}.

\reviseadd{We then investigate the necessity of information transfer at layer 15 and onward, replacing the attention components with full-peeking using ABP. In particular, we prohibit all non-last tokens to attend to any other token except for itself and the BOS token, while allowing the last token to still attend to every previous token. This operation drastically changes the amount of information flow across tokens. However, we observe no performance degradation, suggesting that non-last tokens do not need to receive information from elsewhere and only need to send information to the last one.}


Finally, we incrementally remove the last token full-peeking and restrict it to self-peeking \reviseadd{(where the last token attends only to itself and the BOS token)} from the back. Fig.~\ref{fig:discovery} (bottom) tracks the faithfulness when we modify layer $\ltrans$ to from 0 to 17 (both inclusive) with the layers $L \ge \ltrans + \lwait$ being self-peeking. We find that performance stays high as long as $\ltrans \ge 2$, or in other words, the last token can access other tokens at least layer 15 and 16.

Thus, we have identified a subgraph for Llama-3-8B on the task of $A+B+C$ that retains almost full performance, with 14 CAMA layers and 2 information transfer layers followed by last token self-computation. We call this \modelname$_\mathrm{llama}$. In subsequent sections, we study its performance on other tasks for both Llama-3-8B and Llama-3.1-8B, as well as analogous subgraphs in Pythia and GPT-J. 

\subsection{\modelname{}$_\mathrm{llama}$ Subgraph Performance}\label{sec:subgraph-performance}

\begin{table}[!t]
    \centering
    \resizebox{\columnwidth}{!}{
    \begin{tabular}{llr|llr}\toprule
        Operation & Model & Faithfulness & Operation & Model & Faithfulness\\\midrule
        $A+B+C$ & Llama-3-8B & 0.995 & $A+B$ & Llama-3-8B & 0.854 \\
        $A+B-C$ & Llama-3-8B & 0.944 & $A-B$ & Llama-3-8B & 0.987 \\
        $A-B+C$ & Llama-3-8B & 0.312 & $A\times B$ & Llama-3-8B & 0.710\\
        $A-B-C$ & Llama-3-8B & 0.995 & $A\div B$ & Llama-3-8B & 0.887 \\\midrule
        $A+B+C$ & Llama-3.1-8B & 0.995 & $A+B$ & Llama-3.1-8B & 0.771 \\
        $A+B-C$ & Llama-3.1-8B & 0.974 & $A-B$ & Llama-3.1-8B & 0.889 \\
        $A-B+C$ & Llama-3.1-8B & 0.967 & $A\times B$ & Llama-3.1-8B & 0.503\\
        $A-B-C$ & Llama-3.1-8B & 0.983 & $A\div B$ & Llama-3.1-8B & 0.779 \\\bottomrule
    \end{tabular}
    }
    \caption{\modelname{}$_\mathrm{llama}$ circuit faithfulness for different math operations, on both Llama-3-8B and Llama-3.1-8B.}
    \label{tab:af1-accuracy-llama}
\end{table}

Tab.~\ref{tab:af1-accuracy-llama} presents faithfulness of the same \modelname$_\mathrm{llama}$ subgraph on eight tasks across both Llama models. Despite the sparsity of connections, this subgraph demonstrates high performance in many tasks. The most notable exception is on $A-B+C$ \reviseadd{for Llama-3-8B}, with faithfulness just over 0.3. Considering that the original model accuracy is below 0.3 (Tab.~\ref{tab:raw-accuracy}), the low subgraph faithfulness and low model accuracy may be both caused by inconsistent problem-solving logic used by the model on these inputs. \reviseadd{By contrast, when Llama-3.1-8B achieves a high raw accuracy on this task, \modelname$_\mathrm{llama}$ attains similarly high subgraph faithfulness.}

\begin{figure}[!t]
    \centering
    \vspace{0.175in}
    \includegraphics[width=0.98\linewidth]{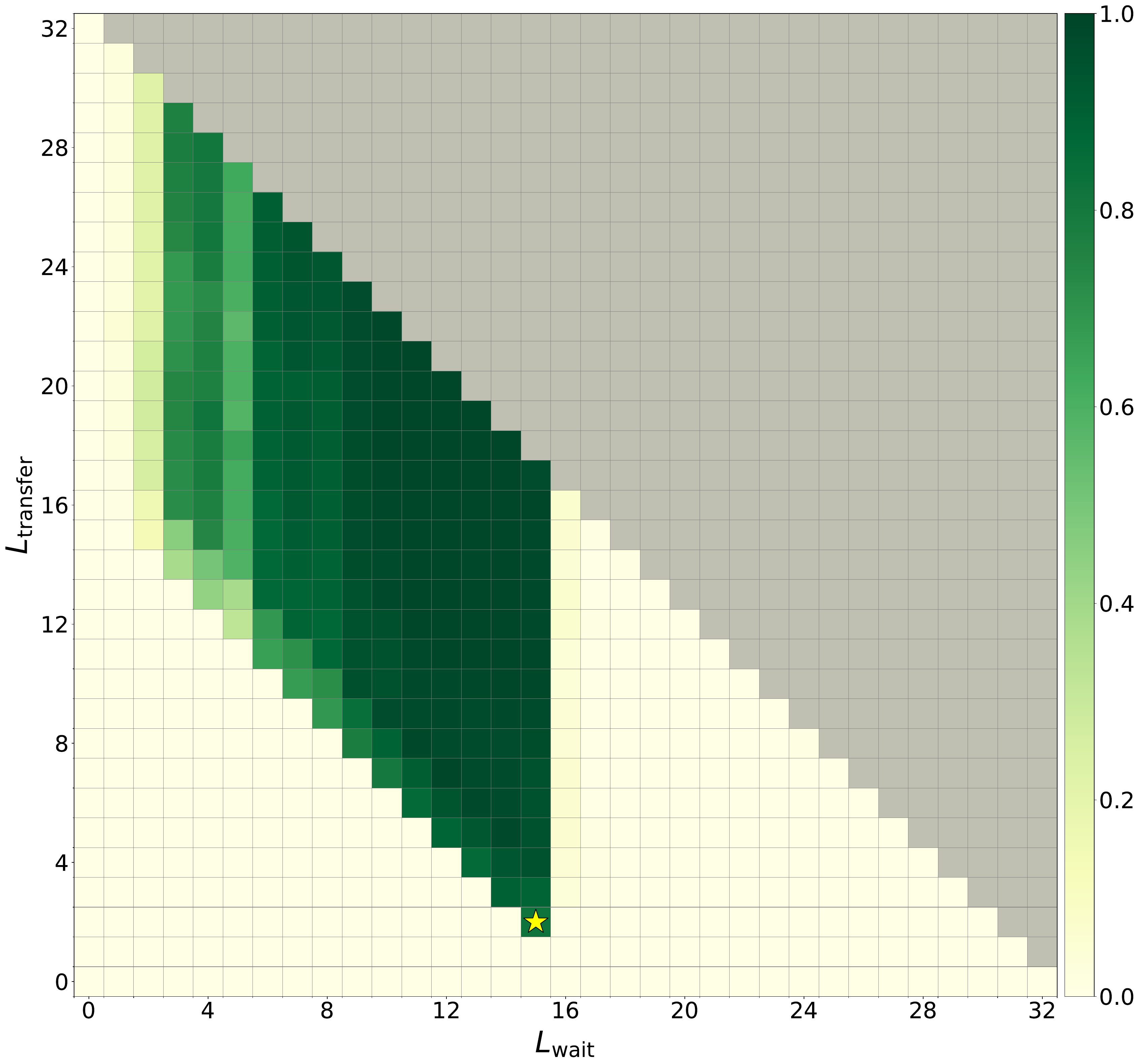}
    \caption{Faithfulness of different \modelname{} configurations over $\ltrans \in [0, 32], \lwait \in [0, 32]$ for $A + B + C$ task on Llama-3-8B model. The minimal subgraph \modelname$_\mathrm{llama}$ is marked with a yellow star. Two conditions are necessary for preserving model accuracy: (1) Waiting can never occur past layer \reviseadd{index} 14 ($\lwait \leq 15$); (2) Information transfer must cover layer 16 (i.e., $\lwait+\ltrans\geq 17$).}
    \label{fig:heatmap-plus-plus}
\end{figure}

Fig.~\ref{fig:heatmap-plus-plus} plots the faithfulness of the subgraph with various $\lwait$ and $\ltrans$ values (with remaining layers implementing last-token self-peeking), for Llama-3-8B on the $A+B+C$ task. We observe the following necessary conditions for high performance: (1) waiting must stop at layer 14 \emph{at the latest} (i.e., $\lwait\leq 15$), and (2) the subsequent information transfer must \reviseadd{be at least two layers long} (i.e., $\lwait+\ltrans\geq 17$). 

Additionally, replacing CAMA layers with information transfer layers too early (i.e., at or before layer 6) actually hurt the performance. Since in an information transfer layer, each non-last token can only attend to itself, without access to task-general context as in CAMA, its ability to perform task-general computation is limited, which proves to be important especially for early layers. 
\reviseadd{A full grid of $L_{\text{wait}}$, $L_{\text{transfer}}$ for all other tasks for both Llama-3-8B and Llama-3.1-8B can be found in App.~\ref{app:full-results}.}






\subsection{Necessity of Information Transfer Layers}
\label{sec:necessity-of-transfer-layers}

\begin{figure}
    \centering
    \includegraphics[width=0.98\linewidth]{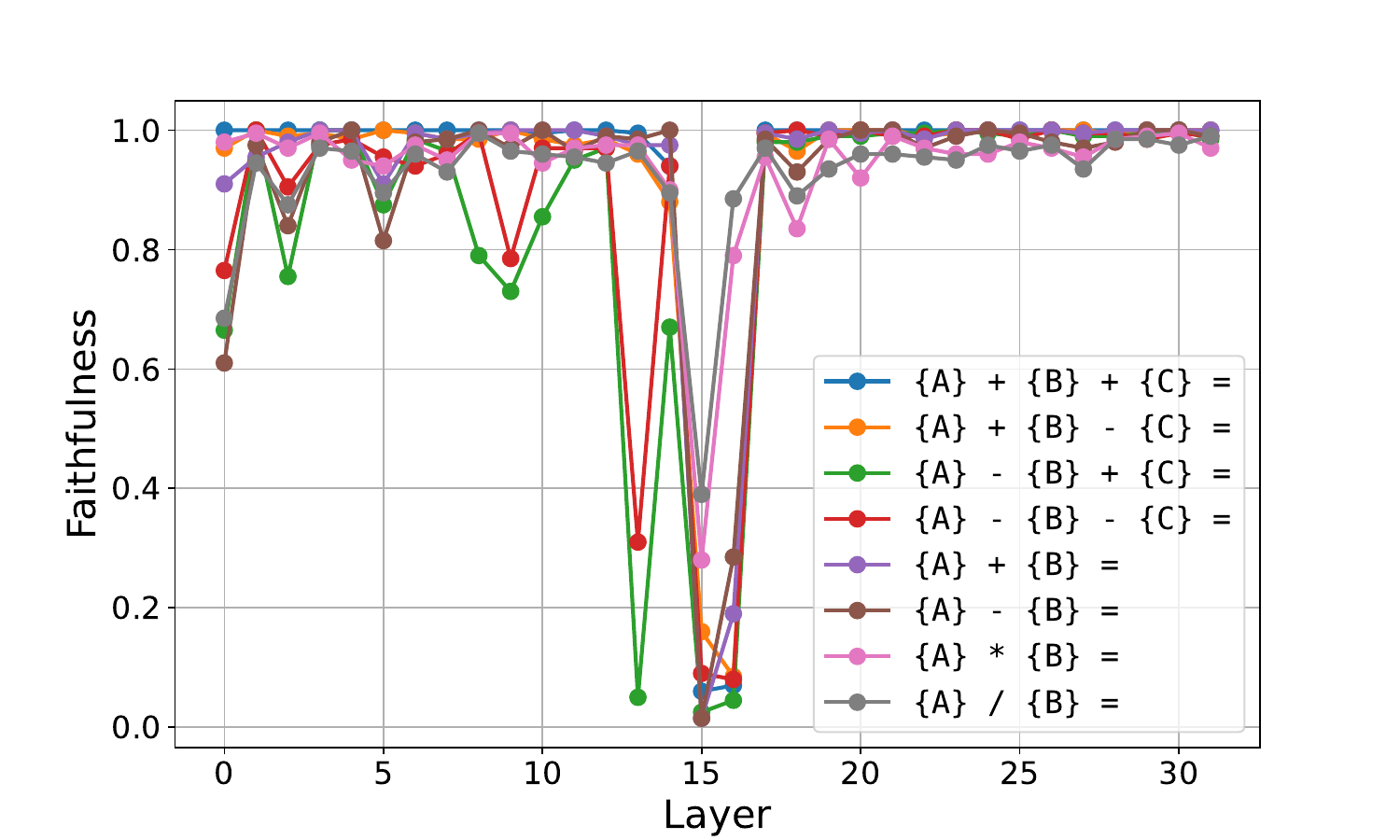}
    \caption{Faithfulness of the full Llama-3-8B but with the attention from the last token to every other non-BOS token removed individually in each layer.}
    \label{fig:attention-removal-faithfulness}
\end{figure}

\reviseadd{To test the necessity of the information transfer layers identified (layers 15 \emph{and} 16) in Llama-3-8B}, we run the full model computation but, for each \textit{single} layer, remove the attentions from the last token to every other non-BOS tokens, while keeping all other attention connections. The resulting model performances are shown in Fig.~\ref{fig:attention-removal-faithfulness}.

Despite removing only $T-2$ out of $T\cdot L\cdot N(N-1)/2$ connections, the effect can be drastic. Without an exception, removing layer 15 attentions to a large performance drop on all tasks, while removing layer 16 attentions affects all but two tasks. This provides further evidence that these two specific attention layers are indeed critical for information transfer. This information, combined with the model's inability to wait past layer 14, suggests special importance of information transfer in layer 15 and 16, agreeing with prior work \citep{nikankin2024arithmetic} that identifies their attention heads being responsible for two-operand arithmetic. As a side note, layer 13 attention is apparently also important for $A-B\pm C$ tasks, although \modelname$_\mathrm{llama}$ (with layer 13 being in the CAMA waiting stage) performs extremely well on $A-B-C$. We leave this investigation to future work.

\reviseadd{In \modelname{}$_\mathrm{llama}$, since only layer 15 and 16 participate in information transfer, we further investigate the importance of each attention head. To this end, we identify the minimal set of heads that preserves the model's performance, through an iterative process as follows (in Llama-3-8B). In Llama, each layer has 32 heads, leading to 64 heads in total for layers 15 and 16. We start with the full \modelname{}$_\mathrm{llama}$ subgraph and remove attention heads iteratively. At every iteration, for each attention head remaining, we compute the model accuracy without this head, and remove the one with the lowest impact. We repeat this procedure until no head remains.} 

\reviseadd{Tab.~\ref{tab:head-removal-acc} shows the accuracy following head removal. We see that very few heads are actually important for arithmetic computation, with 95\% accuracy preserved after removing nearly 60 heads. Furthermore, we identified some common heads, marked with asterisks. } 



\begin{figure}[!b]
    \centering
    \includegraphics[width=\linewidth]{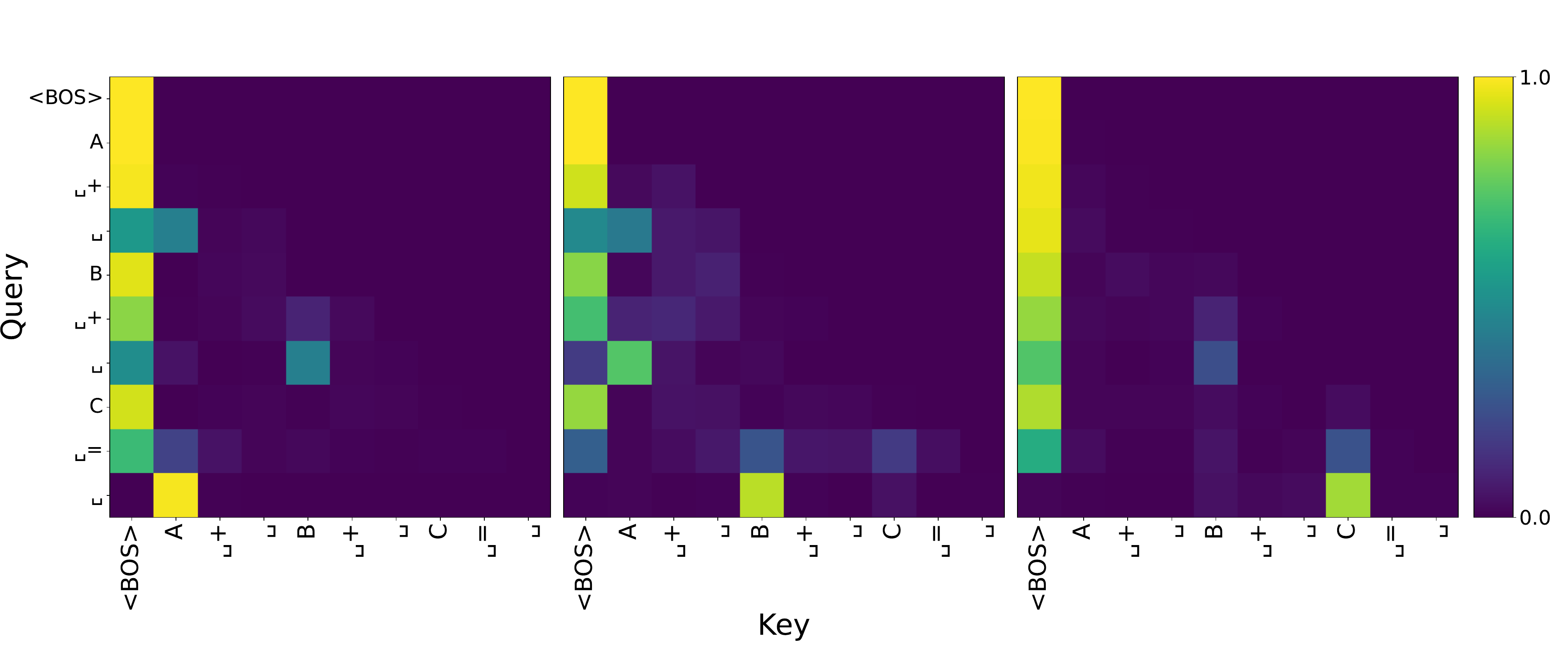}
    \caption{\reviseadd{Attention patterns for the three key operand heads in the $A+B+C$ task (Llama-3-8B): L16H21 (left), L15H13 (middle), L15H3 (right), attending from the last token to the first, second and third operands. Activation values are averaged across 100 prompts.}}
    \label{fig:llama3-add-heads}
\end{figure}

Further attention analysis on these individual heads reveals that some mostly attend to the BOS token, and the remaining heads attend to one of the numerical operands. For example, we show in Fig.~\ref{fig:llama3-add-heads} that heads , L15H13 and L15H3 attend to the first, second, and third operand respectively, for the $A+B+C$ task. This lines up quite neatly with our information transfer hypothesis, with the focused, operand-heavy attention patterns transferring the operand information to the last token. While this analysis is performed on the \modelname{}$_\mathrm{llama}$ subgraph, we observe highly similar results for full Llama-3-8B model as well (App.~\ref{app:full-model-layer-analysis}), suggesting that \modelname{}$_\mathrm{llama}$ captures the ``essence'' of computation.

\begin{table}[!t]
    \centering
    \vspace{0.1in}
    \resizebox{\columnwidth}{!}{
    \begin{tabular}{lr|lr}
        \toprule
        \multicolumn{2}{c|}{$A+B+C$} & \multicolumn{2}{c}{$A+B-C$} \\
        \cmidrule(lr){1-2} \cmidrule(lr){3-4}
        Heads Removed & Accuracy & Heads Removed & Accuracy \\
        \midrule
        59 Least Important & 95.5\% & 56 Least Important & 95.0\% \\
        L15H31 & 93.0\% & L16H20 & 93.5\% \\
        L16H1 * & 64.5\% & L15H6 & 91.5\% \\
        L15H13 * & 8.5\% & L15H3 * & 83.5\% \\
        L15H3 * & 1.5\% & L16H1 * & 46.0\% \\
        L16H21 * & 0.5\% & L16H2 & 30.5\% \\
        - & - & L16H3 & 3.5\% \\
        - & - & L16H21 * & 1.5\% \\
        - & - & L15H13 * & 0.5\% \\
        \bottomrule
    \end{tabular}
    }
    \caption{Preserved accuracy across \textit{cumulative} head removals in layers 15 and 16 on \modelname{}$_\mathrm{llama}$ for $A+B+C$ and $A+B-C$. Asterisks denote heads shared across both tasks.}
    \label{tab:head-removal-acc}
\end{table}

\subsection{Internal Representation Analysis}\label{sec:logitlens}

We probe the final token's residual stream at each layer using \textbf{logit lens analysis}, which projects the residual stream values after each MLP layer to the vocabulary space with the unembedding matrix \citep{nostalgebraist2020blog}. Additionally, we also apply this technique to each attention head. To observe the effect of the \modelname{}$_\mathrm{llama}$ modification, we compare the results on both the full \reviseadd{Llama-3-8B} model and its \modelname{}$_\mathrm{llama}$ subgraph.

\begin{figure}[!t]
    \centering
    \includegraphics[width=1\linewidth]{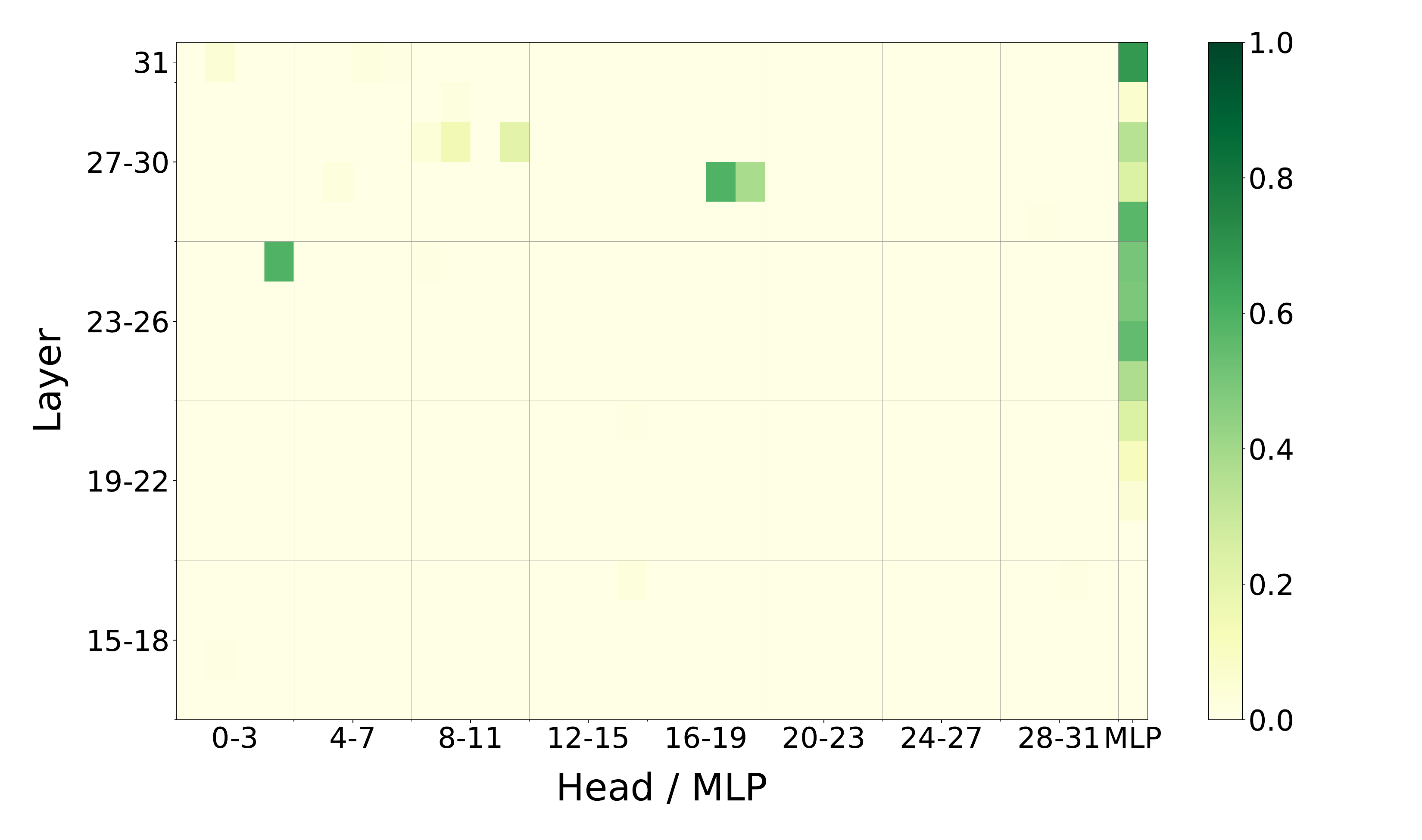}
    \includegraphics[width=1\linewidth]{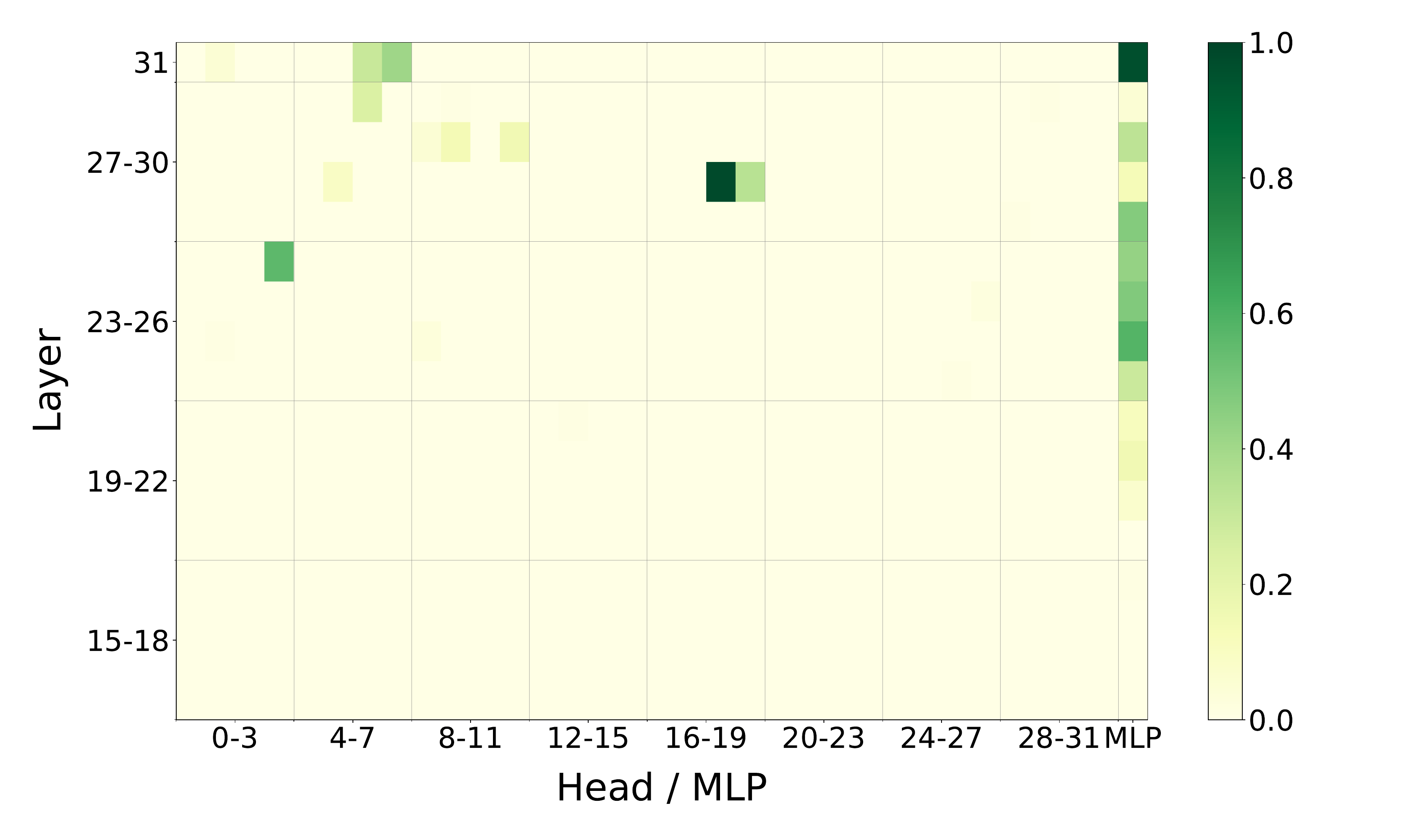}
    \caption{Logit lens top-3 accuracy of for each attention head and the MLP layer on the full \reviseadd{Llama-3-8B model} (top) vs. its \modelname{} subgraph (bottom) on Llama-3-8B.}
    \label{fig:logit-lens}
\end{figure}

Fig.~\ref{fig:logit-lens} depicts the logit lens top-3 accuracy for the full Llama-8-8B (top) and \modelname{}$_\mathrm{llama}$, defined as the fraction of inputs for which the correct answer appears among the top-3 vocab logits, at each attention head as well as the MLP layer. We see high accuracy emerging clearly around layer 24 from the logit lens analysis in Fig.~\ref{fig:logit-lens}. The similarity of the two plots suggest \modelname{}'s captures almost the full prediction power and underlying mechanisms as the full model, instead of finding an alternative computational pathway or a ``hacky shortcut'' to solve the arithmetic. 

L28H18 and L26H3 are the only attention heads that consistently predict the correct answer, even with the answer appearing in the MLP components at earlier layers. However, attention visualization on these heads revealed no more insights, as the pattern shows a common attention sink pattern on the BOS token \reviseadd{(App.~\ref{app:important-attention-heads-visualized}). The logit lens analysis for Llama-3.1-8B yields similar results, as shown in App.~\ref{app:llama-3.1-internal-representation-analysis}. } 



\begin{table*}[!htb]
    \centering
    \renewcommand{\arraystretch}{1.3} 
    \resizebox{\textwidth}{!}{
    \begin{tabular}{rr|rr|l}
        \toprule
        Style & Task & Model Acc. & $\modelname{}_{\text{llama}}$ Faith. & Template and Tokenization \\
        \midrule
        \multirow{2}{*}{Original} 
            & $A + B$ & 0.962 & 0.854 & \colora{<BOS>}\colorb{A}\colora{+}\colorb{B}\colora{=} \\
            & $A - B$ & 0.899 & 0.987 & \colora{<BOS>}\colorb{A}\colora{-}\colorb{B}\colora{=} \\
        \midrule
        \multirow{2}{*}{Verbal Math} 
            & $A + B$ & 1.000 & 1.000 & \colora{<BOS>}\colorb{The}\colora{\textvisiblespace{}sum}\colorb{\textvisiblespace{}of}\colora{\textvisiblespace}\colorb{A}\colora{\textvisiblespace{}and}\colorb{\textvisiblespace}\colora{B}\colorb{\textvisiblespace{}is}\colora{\textvisiblespace} \\
            & $A - B$ & 1.000 & 1.000 & \colora{<BOS>}\colorb{The}\colora{\textvisiblespace{}difference}\colorb{\textvisiblespace{}of}\colora{\textvisiblespace}\colorb{A}\colora{\textvisiblespace{}and}\colorb{\textvisiblespace}\colora{B}\colorb{\textvisiblespace{}is}\colora{\textvisiblespace} \\
        \midrule
        \multirow{2}{*}{Question Answering}  
            & $A + B$ & 0.999 & 1.000 & \colora{<BOS>}\colorb{What}\colora{\textvisiblespace{}is}\colorb{\textvisiblespace{}the}\colora{\textvisiblespace{}sum}\colorb{\textvisiblespace{}of}\colora{\textvisiblespace{}}\colorb{A}\colora{\textvisiblespace{}and}\colorb{\textvisiblespace{}}\colora{B}\colorb{?}\colora{\textvisiblespace{}Answer}\colorb{:}\colora{\textvisiblespace{}} \\
            & $A - B$ & 0.991 & 1.000 & \colora{<BOS>}\colorb{What}\colora{\textvisiblespace{}is}\colorb{\textvisiblespace{}the}\colora{\textvisiblespace{}difference}\colorb{\textvisiblespace{}of}\colora{\textvisiblespace{}}\colorb{A}\colora{\textvisiblespace{}and}\colorb{\textvisiblespace{}}\colora{B}\colorb{?}\colora{\textvisiblespace{}Answer}\colorb{:}\colora{\textvisiblespace{}} \\
        \midrule
        \multirow{2}{*}{Instruction} 
            & $A + B$ & 0.999 & 0.995 & \colora{<BOS>}\colorb{If}\colora{\textvisiblespace{}you}\colorb{\textvisiblespace{}add}\colora{\textvisiblespace{}}\colorb{A}\colora{\textvisiblespace{}to}\colorb{\textvisiblespace{}}\colora{B}\colorb{,}\colora{\textvisiblespace{}you}\colorb{\textvisiblespace{}will}\colora{\textvisiblespace{}get}\colorb{\textvisiblespace{}} \\
            & $A - B$ & 1.000 & 0.905 & \colora{<BOS>}\colorb{If}\colora{\textvisiblespace{}you}\colorb{\textvisiblespace{}subtract}\colora{\textvisiblespace{}}\colorb{A}\colora{\textvisiblespace{}to}\colorb{\textvisiblespace{}}\colora{B}\colorb{,}\colora{\textvisiblespace{}you}\colorb{\textvisiblespace{}will}\colora{\textvisiblespace{}get}\colorb{\textvisiblespace{}} \\
        \midrule
        \multirow{2}{*}{Math Word Problem} 
            & $A + B$ & 0.998 & 0.005 & \colora{<BOS>}\colorb{John}\colora{\textvisiblespace{}has}\colorb{\textvisiblespace{}}\colora{A}\colorb{\textvisiblespace{}cookies}\colora{.}\colorb{\textvisiblespace{}Jane}\colora{\textvisiblespace{}has}\colorb{\textvisiblespace{}}\colora{B}\colorb{\textvisiblespace{}cookies}\colora{.}\colorb{\textvisiblespace{}Together}\colora{\textvisiblespace{}they}\colorb{\textvisiblespace{}have}\colora{\textvisiblespace{}} \\
            & $A - B$ & 0.994 & 0.020 & \colora{<BOS>}\colorb{John}\colora{\textvisiblespace{}has}\colorb{\textvisiblespace{}}\colora{A}\colorb{\textvisiblespace{}cookies}\colora{.}\colorb{\textvisiblespace{}He}\colora{\textvisiblespace{}gave}\colorb{\textvisiblespace{}Jane}\colora{\textvisiblespace{}}\colorb{B}\colora{\textvisiblespace{}cookies}\colorb{.}\colora{\textvisiblespace{}John}\colorb{\textvisiblespace{}now}\colora{\textvisiblespace{}has}\colorb{\textvisiblespace{}} \\
        \midrule
        \multirow{2}{*}{Python Program} 
            & $A + B$ & 0.999 & 0.001 & \colora{<BOS>}\colorb{a}\colora{\textvisiblespace{}=}\colorb{\textvisiblespace{}}\colora{A}\colorb{;}\colora{\textvisiblespace{}b}\colorb{\textvisiblespace{}=}\colora{\textvisiblespace{}}\colorb{B}\colora{;}\colorb{\textvisiblespace{}print}\colora{(a}\colorb{\textvisiblespace{}+}\colora{\textvisiblespace{}b}\colorb{)}\colora{\textvisiblespace{}\#}\colorb{\textvisiblespace{}should}\colora{\textvisiblespace{}print}\colorb{\textvisiblespace{}} \\
            & $A - B$ & 1.000 & 0.020& \colora{<BOS>}\colorb{a}\colora{\textvisiblespace{}=}\colorb{\textvisiblespace{}}\colora{A}\colorb{;}\colora{\textvisiblespace{}b}\colorb{\textvisiblespace{}=}\colora{\textvisiblespace{}}\colorb{B}\colora{;}\colorb{\textvisiblespace{}print}\colora{(a}\colorb{\textvisiblespace{}-}\colora{\textvisiblespace{}b}\colorb{)}\colora{\textvisiblespace{}\#}\colorb{\textvisiblespace{}should}\colora{\textvisiblespace{}print}\colorb{\textvisiblespace{}} \\
        \bottomrule
    \end{tabular}
    }
    \caption{Prompt templates and results for alternative representations of $A + B$ and $A - B$ tasks \reviseadd{on the full Llama-3-8B model and corresponding $\modelname{}_{\text{llama}}$ subgraph}. \texttt{A} and \texttt{B} are replaced with actual numerical values, and everything else is rendered verbatim.}
    \label{tab:other-styles-template-and-result}
\end{table*}

\subsection{Generalized Arithmetic Inputs}
\label{sec:other-styles}
We investigate whether $\modelname{}_{\text{llama}}$ can generalize to other arithmetic forms representing the operations $A+B$ and $A-B$ \reviseadd{on Llama-3-8B}. We consider several input templates such as describing the operation verbally and embedding the operation in word problems or Python code. The concrete templates and resulting model performance are shown in Tab.~\ref{tab:other-styles-template-and-result}. Even for text-based arithmetic prompts, $\modelname{}_{\text{llama}}$ retains considerable accuracy for direct arithmetic tasks without additional semantic contexts. However, it completely fails on tasks requiring semantic understanding such as word problem and Python, suggesting that additional components are needed for other capabilities, like understanding natural language or Python program inputs.

\begin{figure}[!t]
    \centering
    \includegraphics[width=0.93\linewidth]{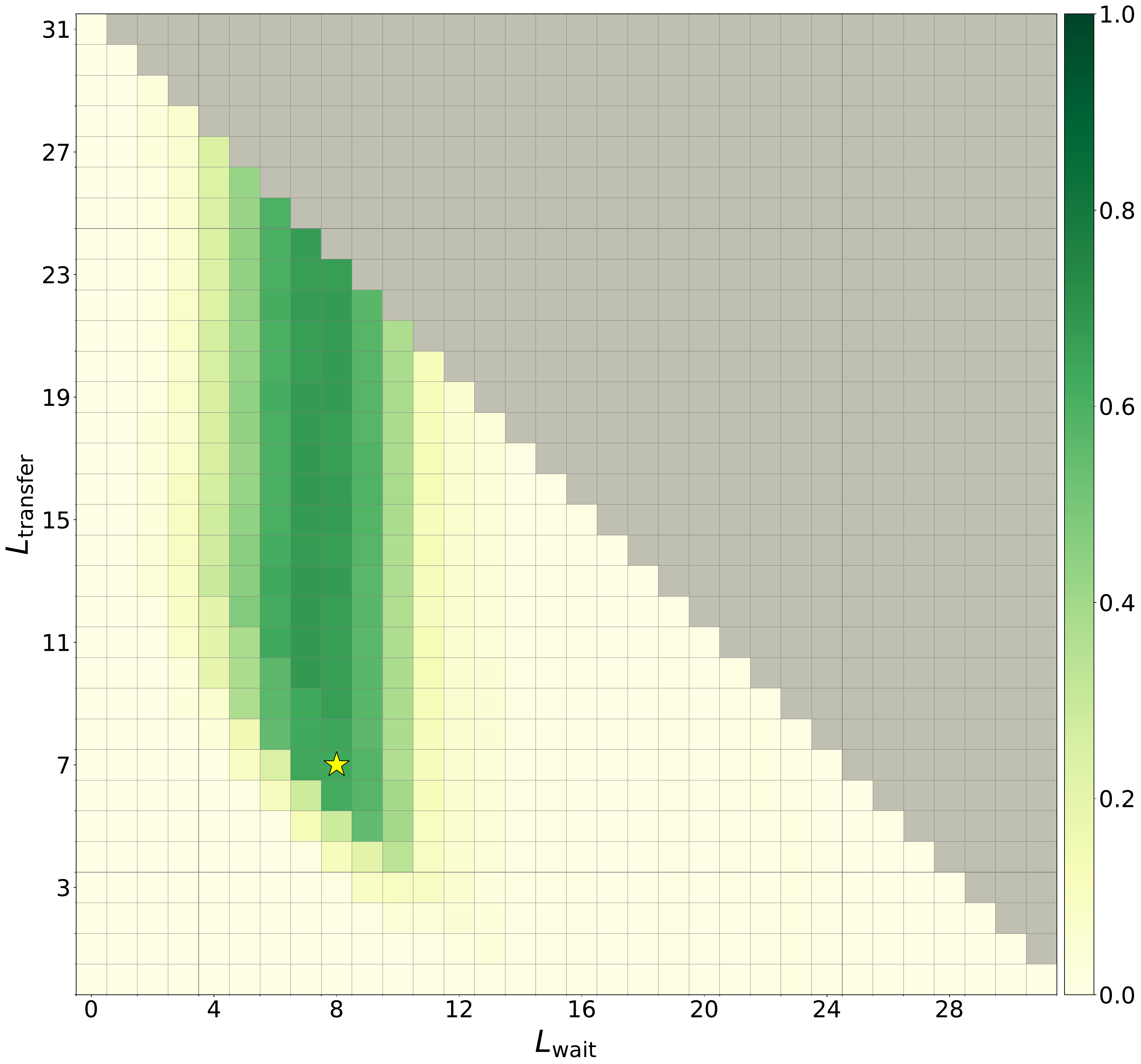}
    \includegraphics[width=0.93\linewidth]{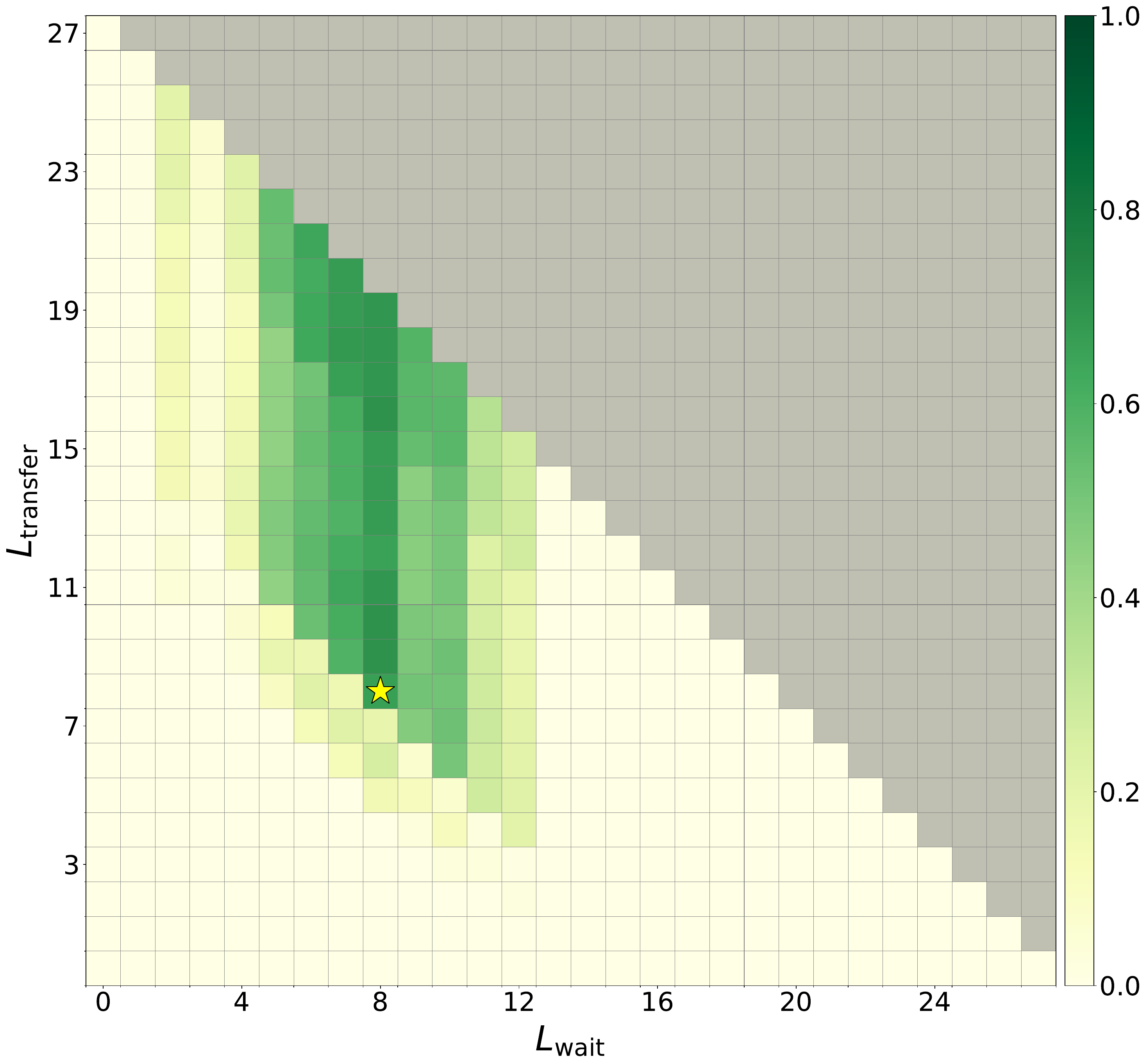}
    \caption{Faithfulness on $A+B$ for different \modelname{} configurations for Pythia (top) and GPT-J (bottom) explored across all possible values of $L_{\text{wait}}$ and $L_{\text{transfer}}$. Tab.~\ref{tab:pythia-gptj-faith} reports the best faithfulness score of the \modelname{} subgraphs, marked with a yellow star in the figure.}
    \label{fig:pythia-gptj-heatmap}
\end{figure}

\subsection{Pythia and GPT-J Models}
In addition to extensively exploring the $\modelname{}_{\text{llama}}$ subgraph, we also applied our experimental procedure using CAMA and ABP to investigate if similar \modelname{} subgraphs exist in other models, namely Pythia and GPT-J. Due to their poor performance on three-operand tasks, we only study two-operand tasks, as done by \citet{nikankin2024arithmetic}.

Fig.~\ref{fig:pythia-gptj-heatmap} plots the faithfulness for different \modelname{} configurations on $A + B$ for Pythia (top) and GPT-J (bottom). There are several notable differences from that for Llama in Fig.~\ref{fig:heatmap-plus-plus}. First, the Pythia and GPT-J grids don't have as definitive of a boundary where performance is nearly completely retained vs. destroyed. Second, the waiting layer must end earlier, with maximum $\lwait$ being around 11 to keep decent performance for both models, suggesting that critical information transfer also happens earlier. Finally, these models require a larger minimum $\ltrans$, suggesting that the efficiency of information transfer is lower in them than in Llama. Based on these factors, we choose $(\lwait, \ltrans)$ to be $(9, 7)$ for \modelname$_\mathrm{Pythia}$ and $(9, 8)$ for \modelname$_\mathrm{GPT{\text -}J}$, marked by stars in Fig.~\ref{fig:pythia-gptj-heatmap}.

\begin{table}[!t]
    \centering
    \vspace{0.07in}
    \resizebox{0.95\columnwidth}{!}{
    \begin{tabular}{lrrrr}
        \toprule
        Model & $A+B$ & $A-B$ & $A\times B$ & $A\div B$ \\
        \midrule
        \modelname{}$_\mathrm{Pythia}$ & 0.620 & 0.551 & 0.780 & 0.490 \\
        \modelname{}$_\mathrm{GPT{\text -}J}$ & 0.647 & 0.506 & 0.794 & 0.440\\
        \bottomrule
    \end{tabular}
    }
    \caption{Faithfulness of $\modelname{}_{\text{Pythia}}$ and $\modelname{}_{\text{GPT-J}}$ for two-operand arithmetic operations. 
    }
    \label{tab:pythia-gptj-faith}
\end{table}

The faithfulness of these two \modelname{} subgraphs for all two-operand tasks are reported in Tab.~\ref{tab:pythia-gptj-faith}. Even though these numbers are usually significantly lower than the Llama counterpart in Tab.~\ref{tab:af1-accuracy-llama}, we see that they still often recover more than half of the original model's accuracy.

\subsection{Alternative CAMA Designs}\label{sec:alternative-cama-abp-designs}

In addition to CAMA, we evaluated several alternative waiting mechanisms in \modelname$_\mathrm{llama}$. \textbf{Direct embedding copy (DEC)} simply uses the original embedding $x_t^{(0)}$ for $\tilde x_t^{(\lwait)}$ in Eq.~\ref{eq:cama-definition}.  \textbf{Random token mean ablation (RTMA)} uses completely random tokens in Eq.~\ref{eq:cama-definition} rather than those drawn from the (context-aware) conditional distribution. \textbf{Self-peek as waiting (SPAW)} uses ABP in every waiting layer with $K_t=\{1, t\}$ to make each token $\{x_t\}_{t=1}^T$ attend to only itself and the BOS token. \textbf{Isolated forward pass (IFP)} runs each token as a standalone two-token prompt (with BOS prepended) for $\lwait$ layers and takes the final representation as $\tilde x_t^{(\lwait)}$ in Eq.~\ref{eq:cama-definition}. \reviseadd{App.~\ref{app:alternative-cama-abp-designs-expanded} contains more detailed explanations of each method.} 

None of the others achieve any non-zero faithfulness. We attribute these failures to two main shortcomings: DEC and RTMA breaks the model’s in-distribution assumptions by directly using input embedding values or averaging over random tokens; SPAW and IFP omit the background computation needed to encode operand structure, providing only minimal structural cues without actual value information. By contrast, CAMA both maintains in-distribution representations via marginalization and captures the model's evolving background computation through conditional expectation.


\section{Discussion and Conclusion}

In this paper, we explored uncovering the least amount of information transfer and computation that support mental math tasks, including those requiring compositional reasoning. Using two new techniques Context-Aware Mean Ablation (CAMA) and Attention-Based Peeking (ABP), we uncovered a highly sparse, three-stage \modelname{} subgraph which allows computation to only happen in the last token's residual stream with information transferred from other tokens in few layers. This subgraph demonstrates high performance on a wide variety of arithmetic tasks across model. 

The discovery and identification of the \modelname{} subgraph provides significant insight into the computational flow of LLMs, particularly highlighting how minimal information transfer periods and targeted, precise computation suffices for high accuracy in arithmetic tasks. Our experiments reveal that although theoretically, tokens can independently process and transfer information between each other from early layers, in practice, meaningful cross-token computation can be (and is) significantly deferred. This underscores the importance of separating task-general computation (such as token recognition and numerical/structural encoding) from input-specific computation (such as carrying out arithmetic operations).

A key feature for \modelname{} is that all input-specific computation is carried out in the last token position, despite every token possessing the ability to implement its own computation. We hypothesize that the root cause lies in the training paradigm. Regardless of pre-training, supervised finetuning or preference alignment, the model receives token-level signal -- predicting the next token based on the context. This dense signal could make the model fully focused on predicting the next token all the time and thus not able to allocate additional bandwidth for compositional reasoning. To remedy this, custom training signals may be explored that places heavier weight on rewarding more ``important'' tokens, which may potentially leading to the emergence of intermediate token computations. 

Overall, this work contributes to the mechanistic understanding of arithmetic reasoning in LLMs and cross-token computation. In addition, it provides methodological innovations with CAMA and ABP that can serve broader applications beyond arithmetic tasks as well.

\section*{Limitations}

The biggest limitation of this analysis is its dependency on a ``cooperative'' tokenizer that allocates a dedicated single token to represent each number, a common practice for recent studies on LLM arithmetic \citep[e.g.,][]{nikankin2024arithmetic}. While the tokenizers for Llama, Pythia and GPT-J have this property, notably exceptions include Qwen \citep{yang2025qwen3} and Gemma \citep{team2024gemma}, which split numbers into individual digit tokens. Studying these models requires extensions of CAMA and ABP to handle multi-token numbers, which we leave to future work.

Additionally, the task scope is limited primarily to arithmetic operations with clearly defined computational boundaries, which can potentially limit generalization to more complex reasoning and more semantically challenging tasks. As demonstrated in Sec.~\ref{sec:other-styles}, \modelname{} notably fails on tasks requiring deeper semantic understanding or context interpretation. Thus, exploring the additional components to \modelname{} that endow models with such capabilities could be useful.

\section*{Acknowledgments}
DR and ZY are funded by the National Science Foundation (Award Number 2311468/2423813).

\bibliography{custom}


\appendix

\begin{figure*}[!t]
  \centering
  \includegraphics[width=0.93\textwidth]{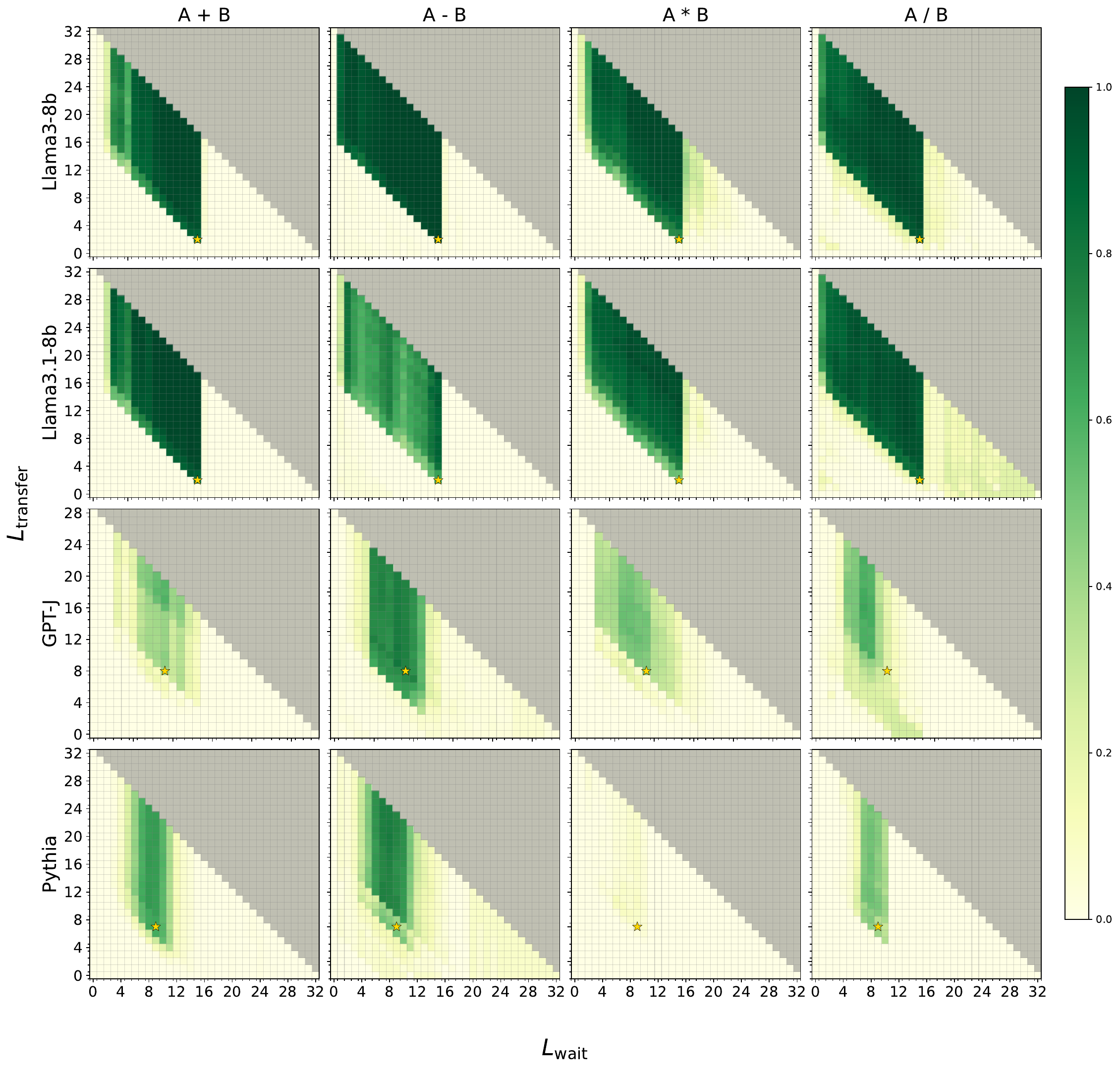}
  \includegraphics[width=0.93\textwidth]{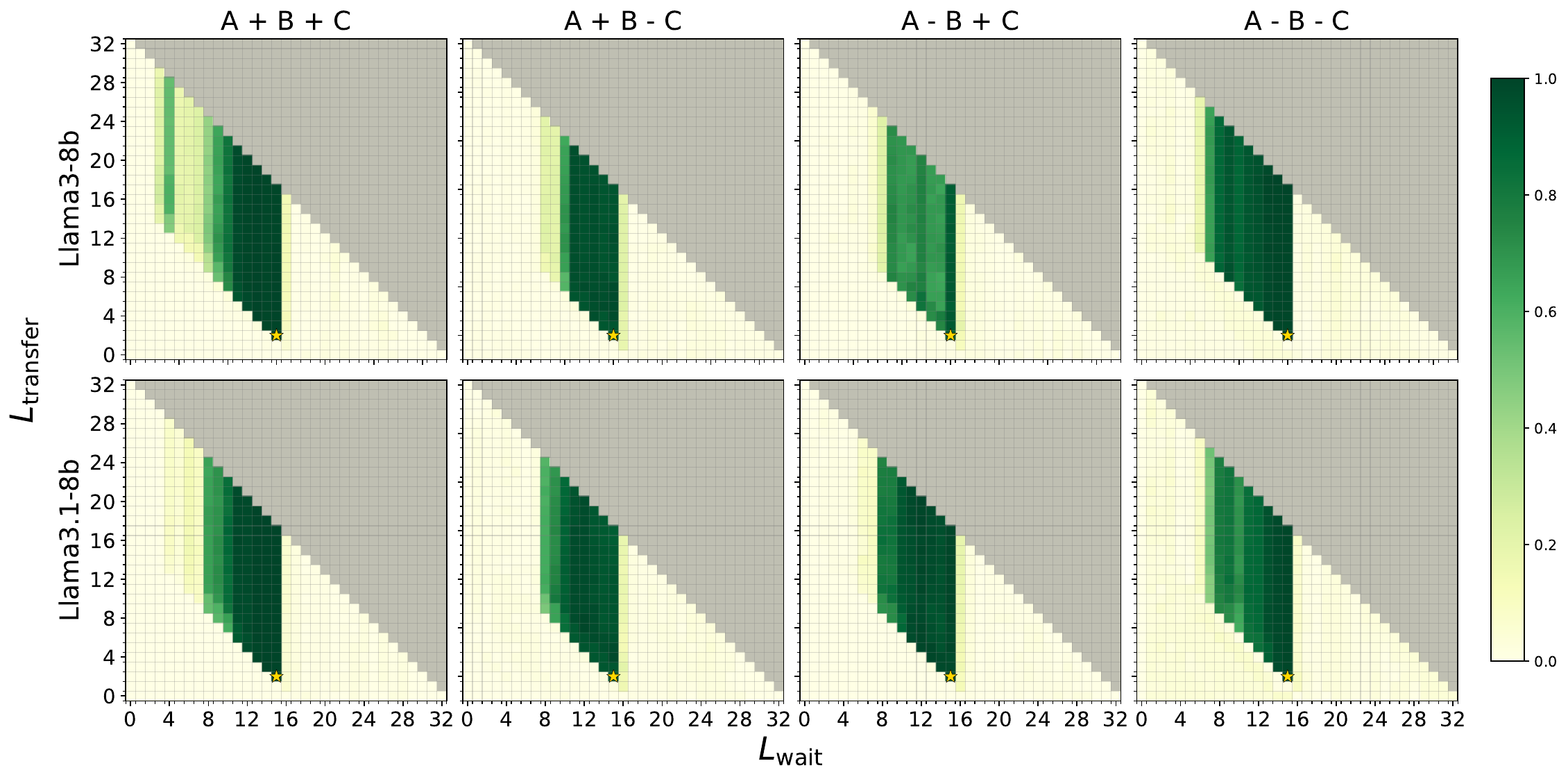}
  \caption{AF1 performance across all models on two- and three-operand tasks.}
  \label{fig:full-plot-23op}
\end{figure*}

\section{Appendix}

\subsection{Implementation Tricks for CAMA}\label{app:cama-expanded}

Recall that CAMA replaces the true representation $x_t^{(\lwait)}$ by the conditional expectation
\begin{align}
\tilde x_t^{(\lwait)}
\;=\;
\mathbb{E}_{\vec{x}'\sim \mathbb P(\vec{x}\mid x_t)}\!\bigl[m(\vec{x}',\,t,\,\lwait)\bigr],
\end{align}
averaging over \emph{in-distribution} prompts that fix the token at position $t$ and vary the rest according to the task distribution. This preserves task-general computation while removing input-specific cross-token information.

Because the self-attention is causal, $x_t^{(\lwait)}$ only depends on tokens at positions $\le t$. Hence, when estimating $\tilde x_t^{(\lwait)}$, it suffices to marginalize over \emph{only prefixes} $x_{1:(t-1)}$ and not consider the value (or even presence) of $x_{(t+1):T}$. \reviseadd{Therefore, when computing the CAMA value for $x_t$, we can safely consider only the first $t$ tokens, sample $x_{1:t-1}$ from $\mathbb P(x_{1:t-1}|x_t)$, and then append the target token value for $x_t$ to the prefix.} 

\reviseadd{Furthermore, there are cases where given a specific value of $x_t$, the probability of the previous token $x_{t-1}$ collapses to a single value, especially if the task distribution is quite restrictive. For example, suppose that the token ``stein'' can only follow ``anken'' in a task containing the word ``Frankenstein''. In this case, we can jointly compute the CAMA values for $x_{t-1}=\text{``anken''}, x_t=\text{``stein''}$ by randomizing their prefix $x_{1:t-2}$ and appending ``ankenstein'' to the input.} 

\reviseadd{More generally, we consider $x_{s:t}$ as one token group if all of $x_{s:t-1}$ are fully determined given a particular value of $x_t$. In this case, computing the CAMA value involves randomizing $x_{1:s-1}$ with the target values of $x_{s:t}$ being appended.}

\reviseadd{In our tasks, since we only vary the operand values and keep everything else in the template fixed (e.g., the BOS token, operators, or spaces), we can form token groups of these fixed tokens and the subsequent operand token. Specifically, we set up the following token groups for the two- and three-operand tasks: }
\begin{align*}
&\groupa{\colora{<BOS>}\colorb{A}}
\groupb{\colora{+}\colorb{B}}
\groupc{\colora{=}}
\\
&\groupa{\colora{<BOS>}\colorb{A}}
\groupb{\colora{\textvisiblespace +}\colorb{\textvisiblespace}\colora{B}}
\groupc{\colorb{\textvisiblespace +}\colora{\textvisiblespace}\colorb{C}}
\groupd{\colora{\textvisiblespace =}\colorb{\textvisiblespace}}
\end{align*}

\reviseadd{Light vs. dark shades represent tokenization, as in Tab.~\ref{tab:task-template}, and each background color represents a particular group. }

\reviseadd{With both the causal attention and the token grouping, computing the CAMA values for a specific group requires the conditional sampling of all values in preceding groups. For example, if we want to compute the CAMA value for the ``\textvisiblespace{}+\textvisiblespace{}C'' group in the three-operand template (green group), we need to sample conditional values of the first two groups (blue and orange). Since all non-operand values are fixed in the template, this amounts to only sampling values of ``A'' and ``B'' to empirically estimate the expectation in Eq.~\ref{eq:cama-definition}.}

\subsection{Performance Grid Visualization}\label{app:full-results}

Fig.~\ref{fig:full-plot-23op} visualize the full grid search across two and three operand arithmetic tasks respectively. These grids are computed across both Llama models, Pythia, and GPT-J. The best AF1 circuit for each model was obtained by empirical observation of all grids for each arithmetic task by each model (and was consistent across both models and all tasks).

\subsection{Analysis of Information Transfer Layers in the Full Llama-3-8B Model}\label{app:full-model-layer-analysis}

\begin{table}[!bt]
    \centering
    \resizebox{\columnwidth}{!}{
    \begin{tabular}{lr|lr}
        \toprule
        \multicolumn{2}{c|}{$A+B+C$} & \multicolumn{2}{c}{$A+B-C$} \\
        \cmidrule(lr){1-2} \cmidrule(lr){3-4}
        Heads Removed & Accuracy & Heads Removed & Accuracy \\
        \midrule
        59 Least Important & 98.8\% & 58 Least Important & 98.6\% \\
        L15H31 & 90.6\% & L16H1 & 88.0\% \\
        L16H1  & 54.8\% & L15H3 & 57.8\% \\
        L15H13 & 6.6\%  & L16H2 & 27.0\% \\
        L16H21 & 2.0\% & L16H3 & 5.6\% \\
        L15H3 & 0.8\%  & L16H21 & 3.2\% \\
        - & - & L15H13 & 2.4\% \\
        \bottomrule
    \end{tabular}
    }
    \caption{Preserved accuracy across \textit{cumulative} head removals in layers 15 and 16 on the full Llama-3-8B model on tasks $A+B-C$, $A+B+C$. All important heads shown here also appear in the study on \modelname{}$_\mathrm{llama}$ as well (Tab.~\ref{tab:head-removal-acc}).}
    \label{tab:head-removal-acc-full-model}
\end{table}

To verify whether the information transfer behavior observed in \modelname{}$_\mathrm{llama}$ reflects the mechanisms of the full Llama-3-8B model, we repeat the head-importance analysis from Sec.~\ref{sec:necessity-of-transfer-layers} on the full Llama 3 8B model, without any \modelname{} modifications. 

Results are presented in Tab.~\ref{tab:head-removal-acc-full-model}. We find that, similar to the AF1 setting, model accuracy remains above 98\% even after 59 of the 64 heads across the two layers. Furthermore, all of the important heads are found in the AF1 subgraph analysis (i.e., appearing in Tab.~\ref{tab:head-removal-acc}). This strong correspondence indicates that the AF1 subgraph captures the same core information transfer heads as the full model, rather than exploiting an alternative mechanism.

These findings reinforce the interpretation that the AF1 subgraph is not a shortcut, but rather a faithful, sparsified version of the model's native computation: in both the subgraph and the full model, a small set of heads in layers 15 and 16 mediate the critical information transfer required for arithmetic reasoning.

\subsection{Important Attention Heads Visualization}\label{app:important-attention-heads-visualized}

\begin{figure}[t]
    \centering
    \includegraphics[width=1\linewidth]{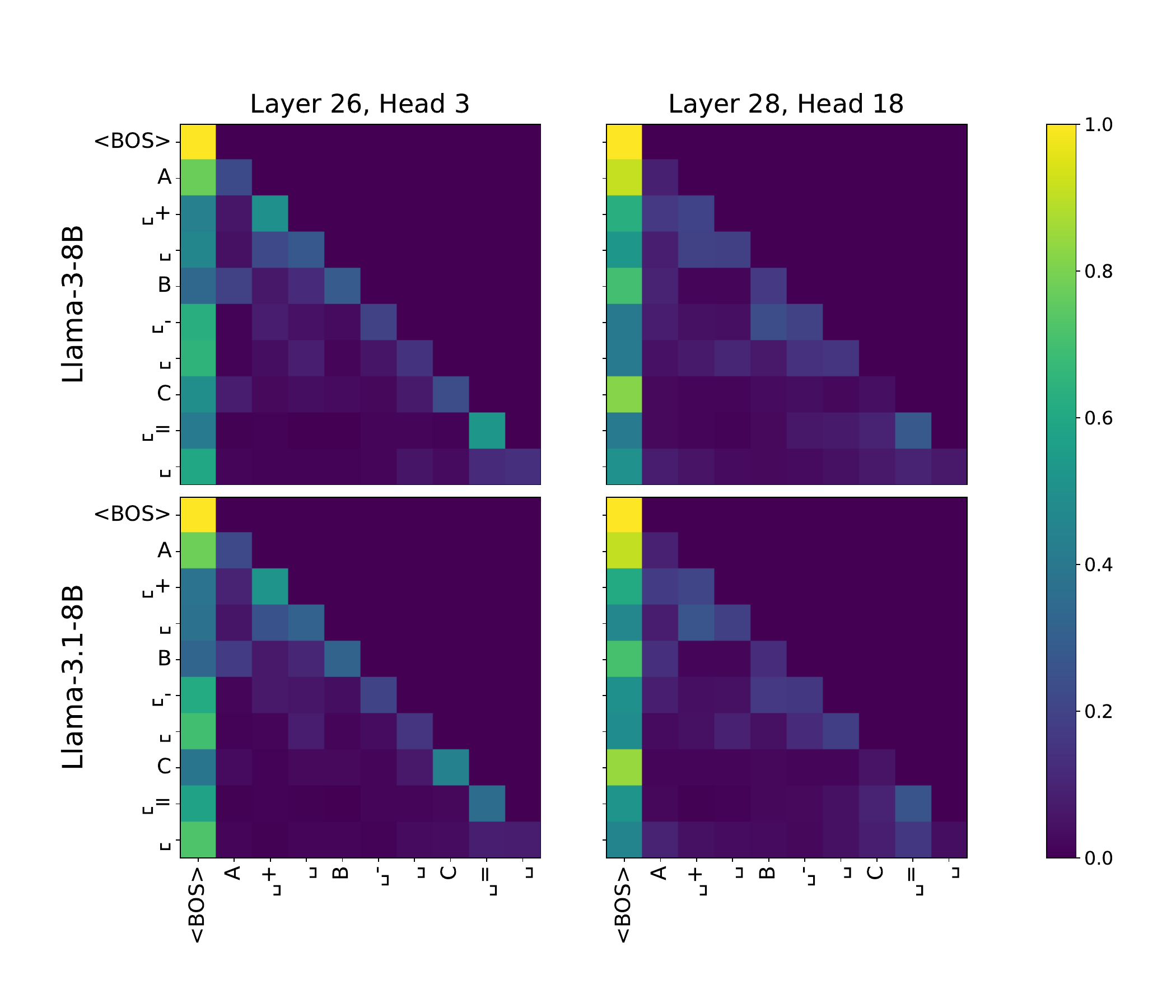}    \caption{Attention pattern for A+B-C, at attention heads at Layer 26, Head 3 and Layer 28, Head 16 on both Llama-3-8B and Llama-3.1-8B. Attention values were averaged over 100 random samples.}
    \label{fig:logit-lens-important-heads}
\end{figure}

Fig.~\ref{fig:logit-lens-important-heads} visualizes the specific attention pattern in the heads which were consistently predicting the right answer (Sec.~\ref{sec:logitlens}). As we can see, there is no clear or interpretable attention patterns in these heads barring a heavy reliance on the <BOS> token, yielding no further insights. Only one prompt is visualized across both key heads, but all tasks demonstrated similarly inconclusive patterns.

\subsection{Llama-3.1-8B Internal Representation Analysis}\label{app:llama-3.1-internal-representation-analysis}

\begin{figure}[t]
    \centering
    \vspace{0.15in}
    \includegraphics[width=1\linewidth]{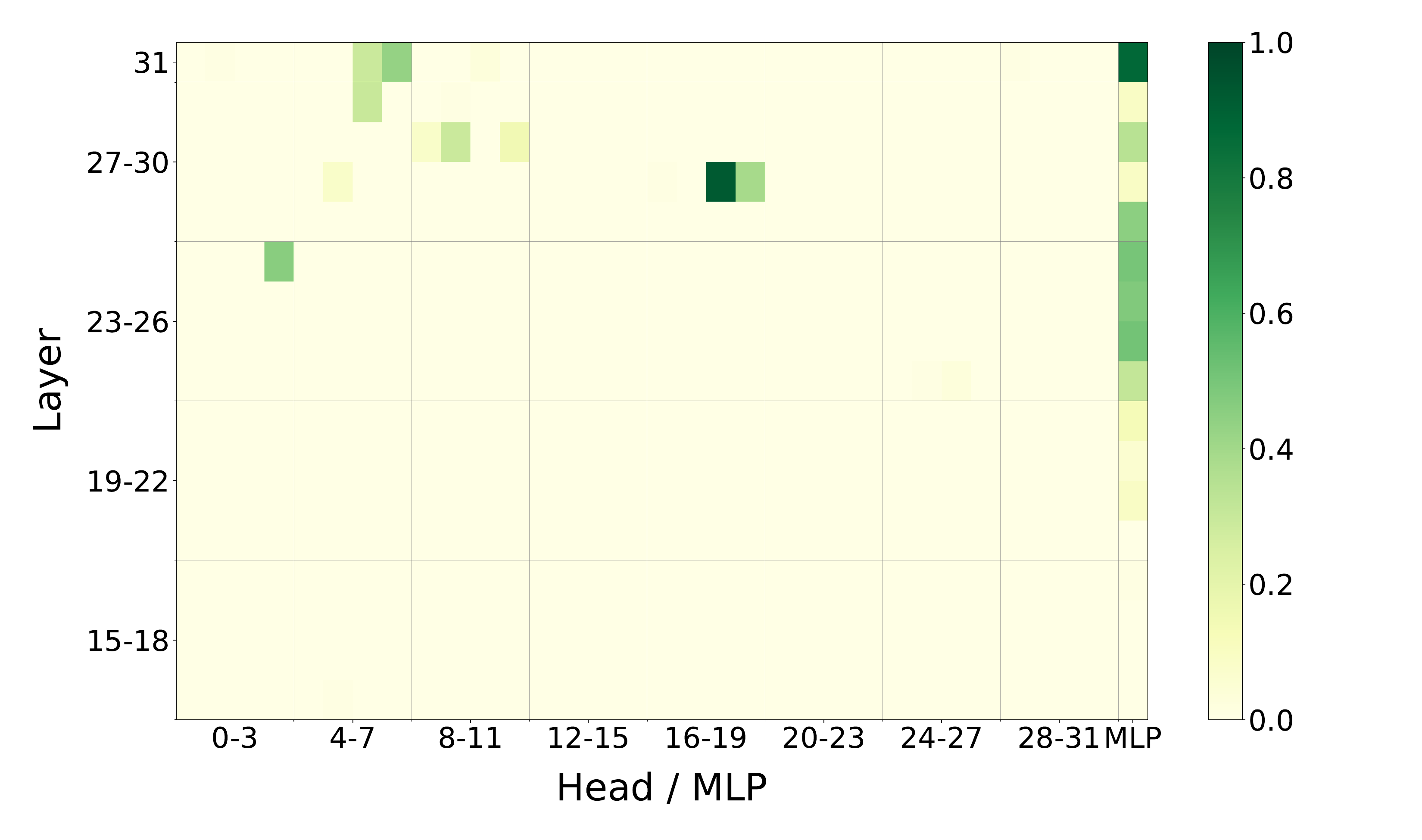}
    \caption{Logit lens top-3 accuracy of for each attention head and the MLP layer on the AF1 subgraph on Llama-3.1-8B.}
    \label{fig:logit-lens-llama3.1}
\end{figure}

Fig.~\ref{fig:logit-lens-llama3.1} depicts the internal representation analysis via logit lens for Llama-3.1-8B. The close correspondence to the Llama 3 plot in Fig.~\ref{fig:logit-lens} suggests that Llama-3.1-8B's arithmetic components remained largely the same despite the improvements on the $A-B-C$ task.

\subsection{Details of Alternative CAMA Designs}
\label{app:alternative-cama-abp-designs-expanded}

\textbf{Direct embedding copy (DEC)} simply uses the original embedding $x_t^{(0)}$ for $\tilde x_t^{(\lwait)}$ in Eq.~\ref{eq:cama-definition}. As mentioned, this likely break's the model's in-distribution assumptions. Even though the information represented at later layers may be the same for these tokens, the embeddings may go through certain transformations which preserve meaning but project through different subspaces of the residual stream.

\textbf{Random token mean ablation (RTMA)} uses completely random tokens in Eq.~\ref{eq:cama-definition} rather than those drawn from the (context-aware) conditional distribution. That is, instead of drawing tokens from the distribution $\mathbb P(\vec x)$, we preserve the prompt length but draw each other token with equal probability from the entire vocab space. This too breaks the in-distribution assumption. While it theoretically captures how the tokens should be projected at any given layer, the lack of conditional awareness ``pollutes'' the context, erasing any arithmetic meaning encoded.

\textbf{Self-peek as waiting (SPAW)} uses ABP in every waiting layer with $K_t=\{1, t\}$ to make each token $\{x_t\}_{t=1}^T$ attend to only itself and the BOS token. In this way, all inter-token computation is erased, but as a result no arithmetic context is encoded through background computation. Additionally, a custom attention mask is directly applied to each token in every waiting layer in every forward pass, causing slowdown issues while CAMA simply substitutes the representations at a key layer.

\textbf{Isolated forward pass (IFP)} runs each token as a standalone two-token prompt (with BOS prepended) for $\lwait$ layers and takes the final representation as $\tilde x_t^{(\lwait)}$ in Eq.~\ref{eq:cama-definition}. Similarly to SPAW, the lack of arithmetic context leads to the inability of the model to perform the necessary information transfer in the key layers.
\end{document}